\documentclass[10pt,twocolumn,letterpaper]{article}
\usepackage[pagenumbers]{cvpr} 
\usepackage{booktabs}
\usepackage{multirow}
\usepackage{graphicx}
\usepackage{array}
\usepackage{gensymb}  
\usepackage{fancyvrb}
\usepackage{lipsum}
\usepackage{xspace}
\usepackage{pifont}

\newcommand{\davgvis}{\ensuremath{\delta_\text{avg}^\text{vis}}\xspace}
\newcommand{\themethod}{VGGT\xspace}
\newcommand{\themethodfull}{Visual Geometry Grounded Transformer\xspace}

\newcommand{\R}{\mathbb{R}}
\newcommand{\bt}{\mathbf{t}}
\newcommand{\by}{\mathbf{y}}
\newcommand{\bp}{\mathbf{p}}

\newcommand{\bq}{\mathbf{q}}
\newcommand{\bg}{\mathbf{g}}
\newcommand{\tok}{\mathrm{t}}

\newcommand{\duster}{DUSt3R\xspace}
\newcommand{\master}{MASt3R\xspace}
\newcommand{\smallnabla}{\scalebox{0.75}{$\nabla$}}

\makeatletter
\renewcommand{\paragraph}{%
    \@startsection{paragraph}{4}%
    {\z@}{-0.5em}{-0.5em}%
    {\normalfont\normalsize\bfseries}%
}
\makeatother

\definecolor{cvprblue}{rgb}{0.21,0.49,0.74}
\usepackage[pagebackref,breaklinks,colorlinks,allcolors=cvprblue]{hyperref}

\def\paperID{13998}
\def\confName{CVPR}
\def\confYear{2025}

\title{\themethod: \themethodfull}

\author{
Jianyuan Wang$^{1,2}$   
\and 
Minghao Chen$^{1,2}$
\and
Nikita Karaev$^{1,2}$    
\and
Andrea Vedaldi$^{1,2}$%
\vspace{0.01cm}
\and 
Christian Rupprecht$^{1}$
\and 
David Novotny$^2$    
\and
\\
$^1$Visual Geometry Group, University of Oxford \hspace{5em} 
$^2$Meta AI 
}

\begin{document}
\twocolumn[{
\maketitle
\begin{center}
\includegraphics[width=\linewidth]{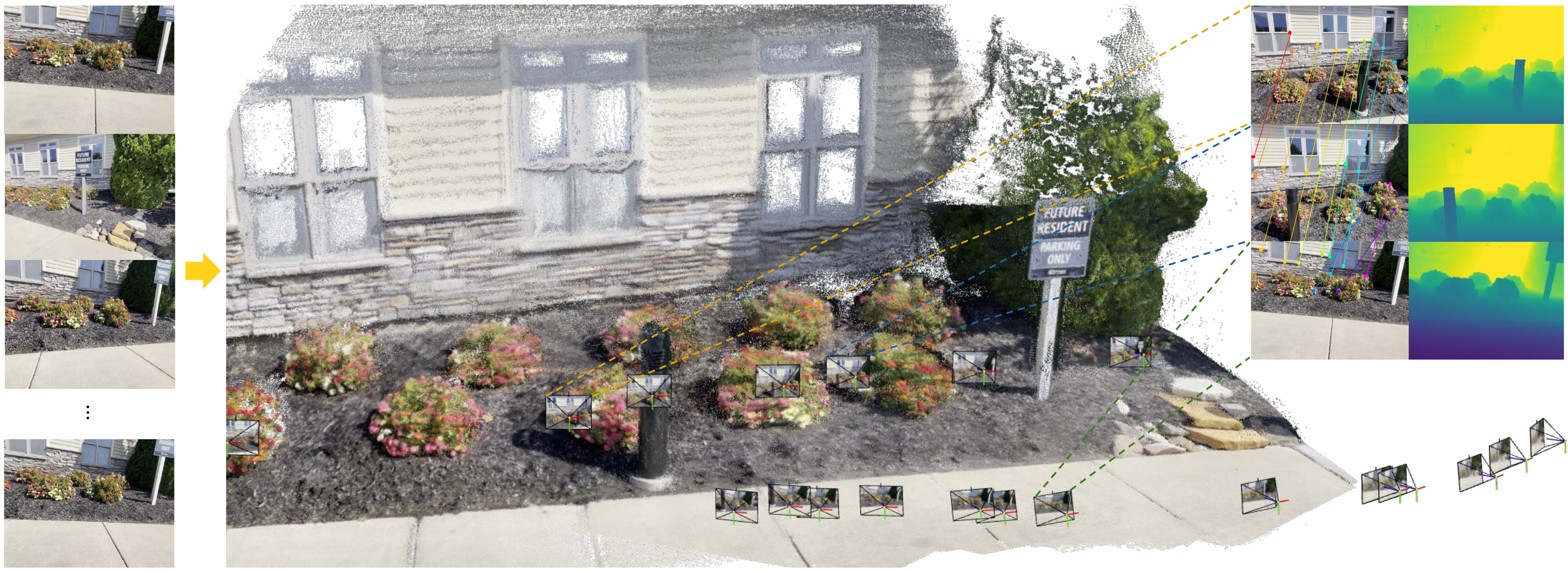}
\end{center}
\vspace{-0.5cm}
\captionsetup{type=figure}
\captionof{figure}{%
\textbf{
\themethod}
is a large feed-forward transformer with minimal 3D-inductive biases trained on a trove of 3D-annotated data.
It accepts up to hundreds of images and predicts cameras, point maps, depth maps, and point tracks for all images at once in less than a second, which often outperforms optimization-based alternatives without further processing. 
}\label{fig:teaser}
\vspace{0.3cm}
}]
\begin{center}
  \large\bfseries Abstract
\end{center}
\textit{%
We present \themethod, a feed-forward neural network that directly infers all key 3D attributes of a scene, including camera parameters, point maps, depth maps, and 3D point tracks, from one, a few, or hundreds of its views.
This approach is a step forward in 3D computer vision, where models have typically been constrained to and specialized for single tasks.
It is also simple and efficient, reconstructing images in under one second, and still outperforming alternatives that require post-processing with visual geometry optimization techniques.  
The network achieves state-of-the-art results in multiple 3D tasks, including camera parameter estimation, multi-view depth estimation, dense point cloud reconstruction, and 3D point tracking. 
We also show that using pretrained \themethod as a feature backbone significantly enhances downstream tasks, such as non-rigid point tracking and feed-forward novel view synthesis. 
Code and models are publicly available at \href{https://github.com/facebookresearch/vggt}{https://github.com/facebookresearch/vggt}. 
%
}


\section{Introduction}%
\label{sec:intro}

We consider the problem of estimating the 3D attributes of a scene, captured in a set of images, utilizing a feed-forward neural network. 
Traditionally, 3D reconstruction has been approached with visual-geometry methods, utilizing iterative optimization techniques like Bundle Adjustment (BA)~\cite{hartley_multiple_2000}.
Machine learning has often played an important complementary role, addressing tasks that cannot be solved by geometry alone, such as feature matching and monocular depth prediction.
The integration has become increasingly tight, and now state-of-the-art Structure-from-Motion (SfM) methods like VGGSfM~\cite{wang24vggsfm:} combine machine learning and visual geometry end-to-end via differentiable BA\@.
Even so, visual geometry \emph{still} plays a major role in 3D reconstruction, which increases complexity and computational cost.

As networks become ever more powerful, we ask if, finally, 3D tasks can be solved \emph{directly} by a neural network, eschewing geometry post-processing almost entirely.
Recent contributions like \duster~\cite{wang24dust3r:} and its evolution \master~\cite{mast3r} have shown promising results in this direction, but these networks can only process two images at once and rely on post-processing to reconstruct more images, fusing pairwise reconstructions.

In this paper, we take a further step towards removing the need to optimize 3D geometry in post-processing.
We do so by introducing \emph{\themethodfull} (\themethod), a feed-forward neural network that performs 3D reconstruction from one, a few, or even hundreds of input views of a scene.
\themethod predicts a full set of 3D attributes, including camera parameters, depth maps,  point maps, and 3D point tracks. 
It does so in a single forward pass, in seconds.
Remarkably, it often outperforms optimization-based alternatives even without further processing.
This is a substantial departure from \duster, \master, or VGGSfM, which still require costly iterative post-optimization to obtain usable results.

We also show that it is unnecessary to design a special network for 3D reconstruction.
Instead, \themethod is based on a fairly standard large transformer~\cite{vaswani17attention}, with no particular 3D or other inductive biases (except for alternating between frame-wise and global attention), but trained on a large number of publicly available datasets with 3D annotations.
\themethod is thus built in the same mold as large models for natural language processing and computer vision, such as GPTs~\cite{yenduri23generative,dubey24the-llama,achiam2023gpt}, CLIP~\cite{radford21learning}, DINO~\cite{caron21emerging,oquab24dinov2:}, and Stable Diffusion~\cite{esser21taming}.
These have emerged as versatile backbones that can be fine-tuned to solve new, specific tasks.
Similarly, we show that the features computed by \themethod can significantly enhance downstream tasks like point tracking in dynamic videos, and novel view synthesis.

There are several recent examples of large 3D neural networks, including DepthAnything~\cite{yang24depth}, MoGe~\cite{wang24moge:}, and LRM~\cite{hong24lrm:}.
However, these models only focus on a single 3D task, such as monocular depth estimation or novel view synthesis.
In contrast, \themethod uses a shared backbone to predict all 3D quantities of interest together.
We demonstrate that \emph{learning} to predict these interrelated 3D attributes enhances overall accuracy despite potential redundancies.
At the same time, we show that, during \emph{inference}, we can derive the point maps from separately predicted depth and camera parameters, obtaining better accuracy compared to directly using the dedicated point map head.

To summarize, we make the following contributions:
(1) We introduce \themethod, a large feed-forward transformer that, given one, a few, or even hundreds of images of a scene, can predict all its key 3D attributes, including camera intrinsics and extrinsics, point maps, depth maps, and 3D point tracks, in seconds.
(2) We demonstrate that \themethod's predictions are directly usable, being highly competitive and usually better than those of state-of-the-art methods that use slow post-processing optimization techniques.
(3) We also show that, when further combined with BA post-processing, \themethod achieves state-of-the-art results across the board, even when compared to methods that specialize in a subset of 3D tasks, often improving quality substantially.

We make our code and models publicly available at \href{https://github.com/facebookresearch/vggt}{https://github.com/facebookresearch/vggt}. 
We believe that this will facilitate further research in this direction and benefit the computer vision community by providing a new foundation for fast, reliable, and versatile 3D reconstruction. 

\begin{figure*}
\centering
\includegraphics[width=\textwidth]{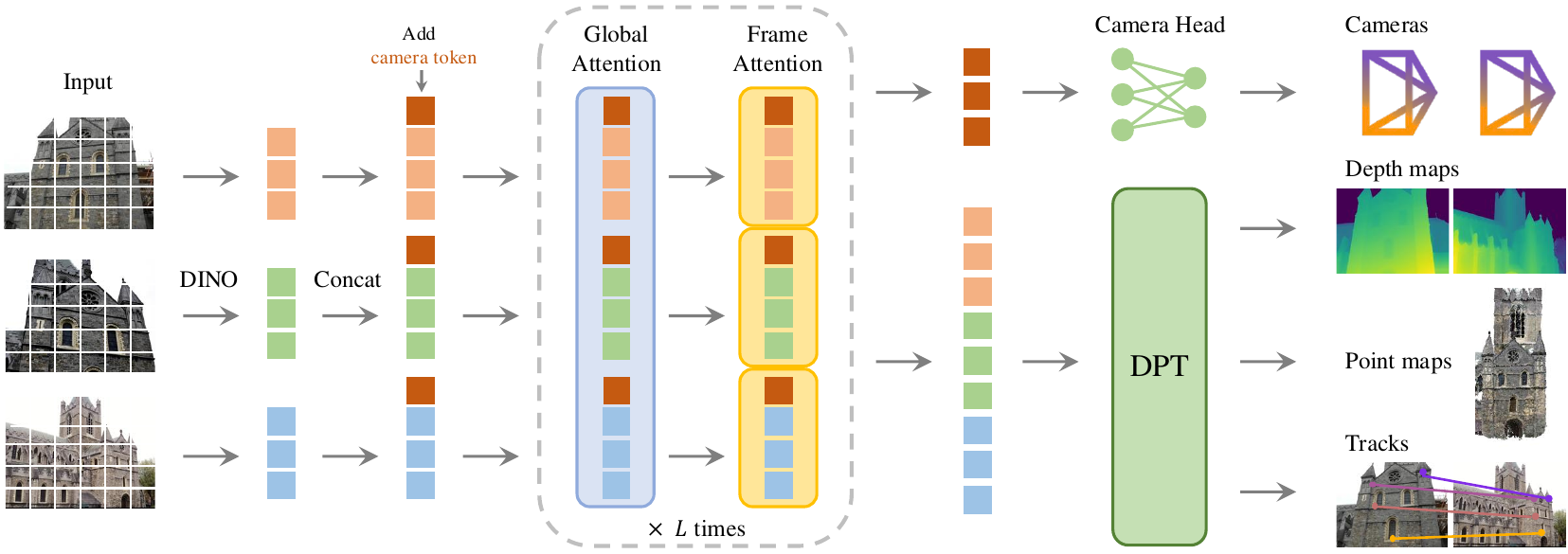}
\caption{\textbf{Architecture Overview.} Our model first patchifies the input images into tokens by DINO, and appends camera tokens for camera prediction.
It then alternates between frame-wise and global self attention layers. 
A camera head makes the final prediction for camera extrinsics and intrinsics, and a DPT~\cite{ranftl21dpt} head for any dense output. 
}%
\label{fig:transformer_architecture_comparison}
\end{figure*}

\section{Related Work}%
\label{sec:rw}

\paragraph{Structure from Motion}
is a classic computer vision problem~\cite{hartley_multiple_2000, ozyecsil2017survey, oliensis2000critique} that involves estimating camera parameters and reconstructing sparse point clouds from a set of images of a static scene captured from different viewpoints.
The traditional SfM pipeline~\cite{snavely2006photo, agarwal2011building, frahm2010building, wu2013towards, schoenberger2016sfm, liu2025robust} consists of multiple stages, including image matching, triangulation, and bundle adjustment.
COLMAP~\cite{schoenberger2016sfm} is the most popular framework based on the traditional pipeline.
In recent years, deep learning has improved many components of the SfM pipeline, with keypoint detection~\cite{yi_lift_2016, detone2018self, dusmanu2019d2, tyszkiewicz2020disk} and image matching~\cite{sarlin2020superglue, chen2021learning, shi2022clustergnn, lindenberger2023lightglue} being two primary areas of focus.
Recent methods~\cite{zhou2017unsupervised, ummenhofer2017demon, tang2018ba, wei2020deepsfm, wang2021deep, teed2018deepv2d, teed2021droid, brachmann2024acezero, smith24flowmap:, wang24vggsfm:} explored end-to-end differentiable SfM, where VGGSfM~\cite{wang24vggsfm:} started to outperform traditional algorithms on challenging phototourism scenarios.

\paragraph{Multi-view Stereo}
aims to densely reconstruct the geometry of a scene from multiple overlapping images, typically assuming known camera parameters, which are often estimated with SfM.
MVS methods can be divided into three categories: traditional handcrafted~\cite{furukawa2015multi, galliani2015massively, schonberger2016pixelwise, Wang_2023_CVPR}, global optimization~\cite{niemeyer2020differentiable, fu2022geo, Wei_2021_ICCV, yariv2020multiview}, and learning-based methods~\cite{yao2018mvsnet, gu2020cascade, ma2022multiview, peng2022rethinking, geomvsnet}.
As in SfM, learning-based MVS approaches have recently seen a lot of progress.
Here, \duster~\cite{wang24dust3r:} and \master~\cite{mast3r} directly estimate aligned dense point clouds from a pair of views, similar to MVS but without requiring camera parameters.
Some concurrent works~\cite{cut3r, tang2024mv, zhang2025flare, yang2025fast3r} explore replacing \duster's test-time optimization with neural networks, though these attempts achieve only suboptimal or comparable performance to \duster. Instead, \themethod outperforms \duster and \master by a large margin.

\paragraph{Tracking-Any-Point}
was first introduced in Particle Video~\citep{sand2008particle} and revived by PIPs~\citep{pips} during the deep learning era, aiming to track points of interest across video sequences including dynamic motions.  
Given a video and some 2D query points, the task is to predict 2D correspondences of these points in all other frames.
TAP-Vid~\citep{tapvid} proposed three benchmarks for this task and a simple baseline method later improved in TAPIR~\citep{doersch2023tapir}.
CoTracker~\citep{karaev2023cotracker, karaev2024cotracker3} utilized correlations between different points to track through occlusions, while DOT~\citep{lemoing2024dense} enabled dense tracking through occlusions.
Recently, TAPTR~\citep{li2024taptr} proposed an end-to-end transformer for this task, and LocoTrack~\citep{cho2024local} extended commonly used pointwise features to nearby regions.
%
%
All of these methods are specialized point trackers.
Here, we demonstrate that \themethod's features yield state-of-the-art tracking performance when coupled with existing point trackers.

\section{Method}%
\label{sec:method}

\begin{figure*}[htbp] 
\centering
\includegraphics[width=\linewidth]{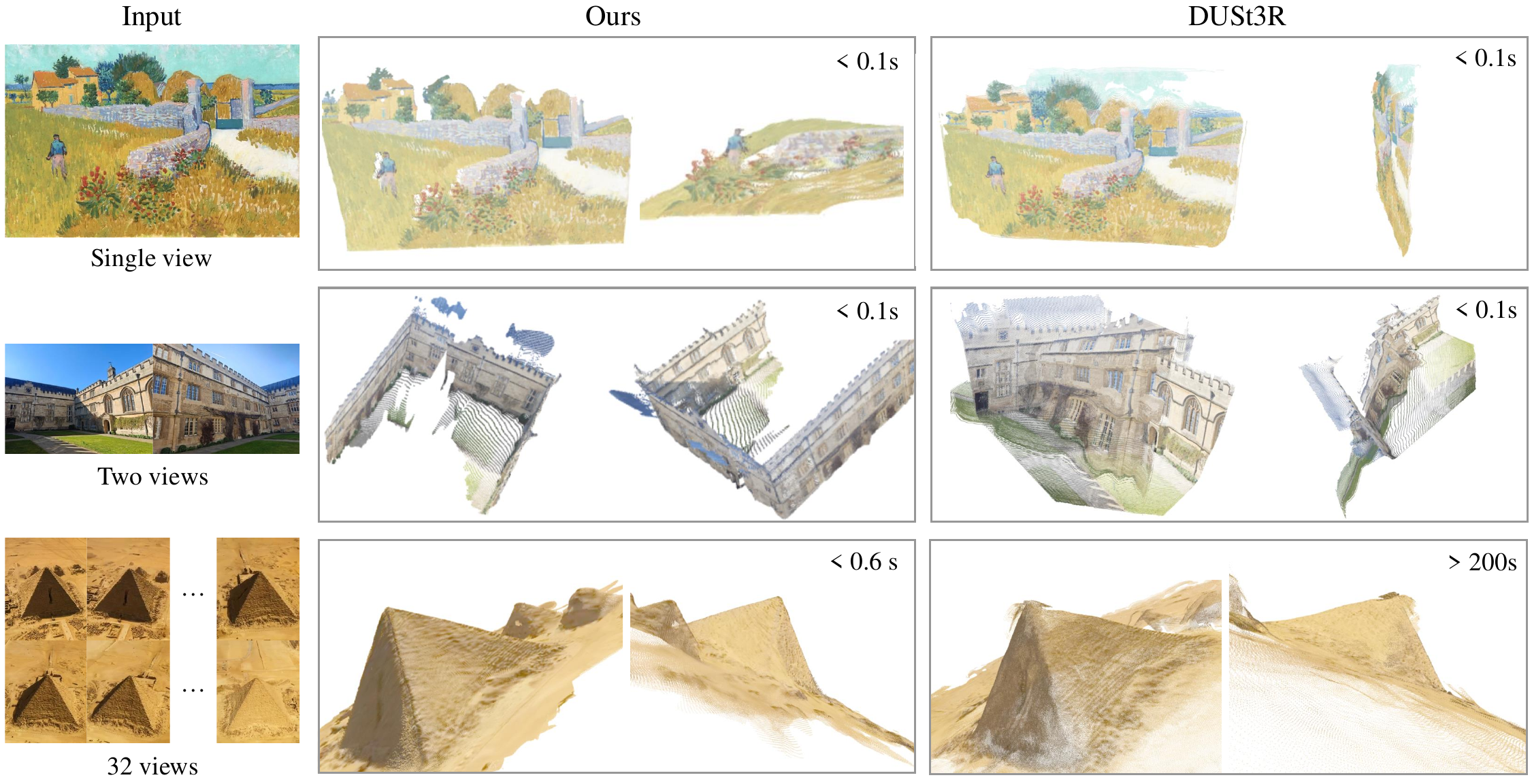}
\caption{\textbf{Qualitative comparison of our predicted 3D points to \duster{} on in-the-wild images.}  
As shown in the top row, our method successfully predicts the geometric structure of an oil painting, while \duster{} predicts a slightly distorted plane.  In the second row, our method correctly recovers a 3D scene from two images with no overlap, while \duster{} fails. 
The third row provides a challenging example with repeated textures,  while our prediction is still high-quality. 
We do not include examples with more than 32 frames, as \duster{} runs out of memory beyond this limit.
}%
\label{fig:pointmap_compare}
\end{figure*}

\begin{figure*}[htbp] 
\centering
\includegraphics[width=\linewidth]{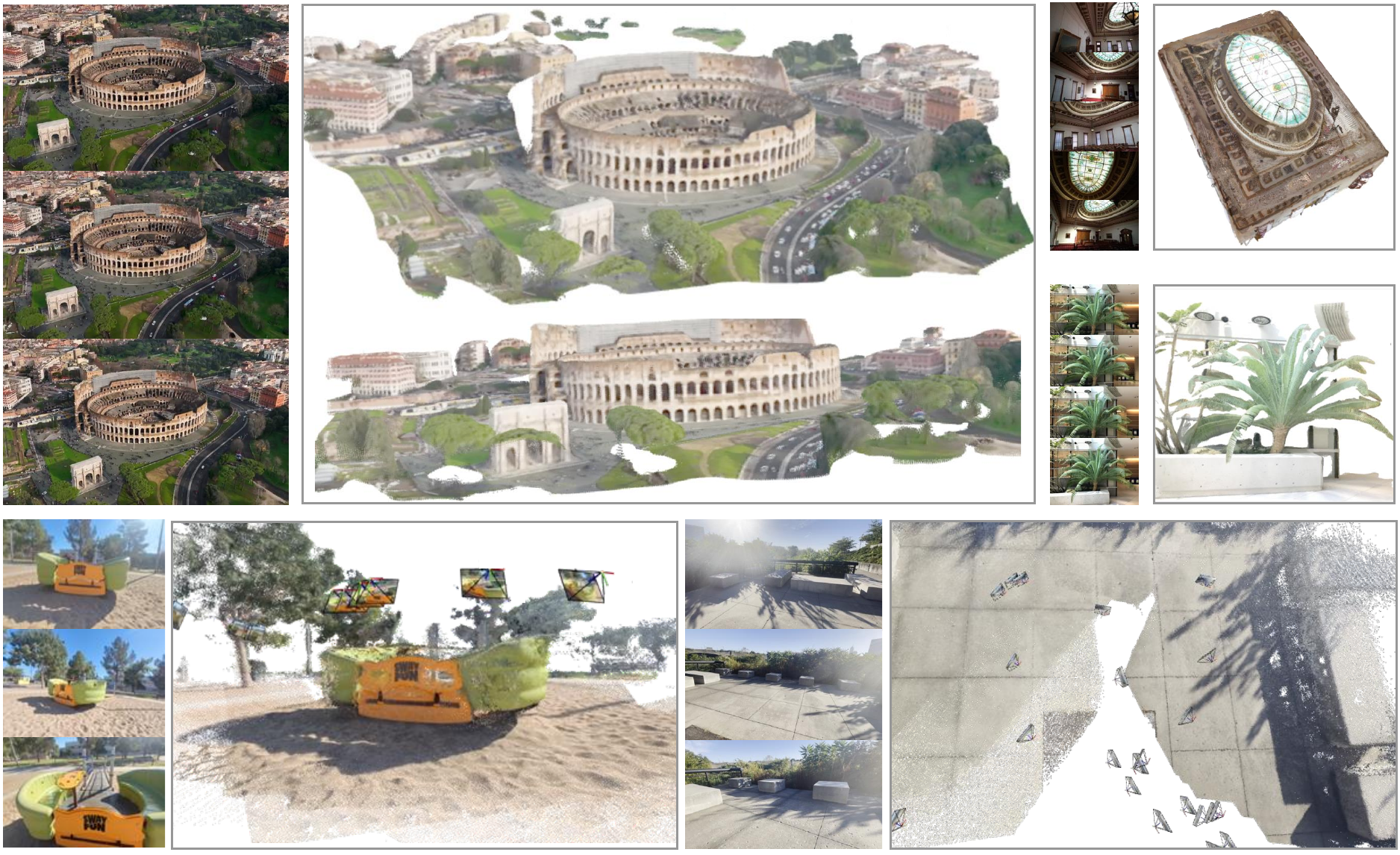}
\caption{\textbf{Additional Visualizations of Point Map Estimation.} Camera frustums illustrate the estimated camera poses.  
Explore our interactive demo for better visualization quality.   
}%
\label{fig:pointmap}
\end{figure*}

We introduce \themethod, a large transformer that ingests a set of images as input and produces a variety of 3D quantities as output.
We start by introducing the problem in \cref{sec:problem_definition}, followed by our architecture in \cref{sec:architecture} and its prediction heads in \cref{sec:prediction_heads}, and finally the training setup in \cref{sec:training_losses}.

\subsection{Problem definition and notation}%
\label{sec:problem_definition}

The input is a sequence $(I_i)_{i=1}^N$ of $N$ RGB images
$
I_i \in \R^{3\times H\times W},
$
observing the same 3D scene.
\themethod's transformer is a function that maps this sequence to a corresponding set of 3D annotations, one per frame:
\begin{equation}
f\left((I_i)_{i=1}^N\right)
=
\left(
\bg_i,
D_i,
P_i,
T_i
\right)_{i=1}^N.
\end{equation}
The transformer thus maps each image $I_i$ to
its camera parameters $\bg_i \in \mathbb{R}^9$ (intrinsics and extrinsics),
its depth map $D_i \in \R^{H\times W}$,
its point map $P_i \in \R^{3\times H\times W}$,
and a grid $T_i \in \R^{C\times H\times W}$ of $C$-dimensional features for point tracking.
We explain next how these are defined.

For the \textbf{camera parameters} $\bg_i$, we use the parametrization from~\cite{wang24vggsfm:} and set
$
\bg = [\mathbf{q}, \mathbf{t}, \mathbf{f}]
$
which is the concatenation of
the rotation quaternion $\bq \in \R^4$,
the translation vector $\bt \in \R^3$,
and the field of view $\mathbf{f} \in \R^2$.
We assume that the camera's principal point is at the image center, which is common in SfM frameworks~\cite{schonberger16structure-from-motion,wang24vggsfm:}. 

We denote the domain of the image $I_i$ with
$
\mathcal{I}(I_i) = \{1, \dots, H\} \times \{1,\dots,W\},
$
\ie, the set of pixel locations.
The \textbf{depth map} $D_i$ associates each pixel location $\by \in \mathcal{I}(I_i)$ with its corresponding depth value $D_i(\by) \in \R^+$, as observed from the $i$-th camera. 
Likewise, the \textbf{point map} $P_i$ associates each pixel with its corresponding 3D scene point $P_i(\by) \in \mathbb{R}^3$.
Importantly, like in \duster~\cite{wang24dust3r:}, the point maps are \emph{viewpoint invariant}, meaning that the 3D points $P_i(\by)$ are defined in the coordinate system of the first camera $\bg_1$, which we take as the world reference frame.

Finally, for \textbf{keypoint tracking}, we follow track-any-point methods such as~\cite{karaev24cotracker,doersch23tapir:}.
Namely, given a fixed query image point $\mathbf{y}_q$ in the query image $I_q$, the network outputs a track
$
\mathcal{T}^\star(\mathbf{y}_q)
=
(
    \mathbf{y}_i
)_{i=1}^N
$
formed by the corresponding 2D points $\mathbf{y}_i \in \R^2$ in all images $I_i$.

Note that the transformer $f$ above does not output the tracks directly but instead features $T_i \in \mathbb{R}^{C\times H \times W}$, which are used for tracking.
The tracking is delegated to a separate module, described in \cref{sec:prediction_heads}, which implements a function
$
\mathcal{T}( (\mathbf{y}_j)_{j=1}^M, (T_i)_{i=1}^N)
= ((\hat{\by}_{j,i})_{i=1}^N)_{j=1}^M.
$
It ingests the query point $\mathbf{y}_q$ and the dense tracking features $T_i$ output by the transformer $f$ and then computes the track.
The two networks $f$ and $\mathcal{T}$ are trained jointly end-to-end.

\paragraph{Order of Predictions.}

The order of the images in the input sequence is arbitrary, except that the first image is chosen as the reference frame.
The network architecture is designed to be permutation equivariant for all but the first frame.

\paragraph{Over-complete Predictions.}

Notably, not all quantities predicted by \themethod are independent.
For example, as shown by \duster~\cite{wang24dust3r:}, the camera parameters $\bg$ can be inferred from the invariant point map $P$, for instance, by solving the Perspective-$n$-Point (PnP) problem~\cite{lepetit2009ep, fischler1981random}.
Furthermore, the depth maps can be deduced from the point map and the camera parameters.
However, as we show in \cref{sec:ablation}, tasking \themethod with explicitly predicting \textit{all} aforementioned quantities during training brings substantial performance gains, even when these are related by closed-form relationships. 
Meanwhile, during inference, it is observed that combining independently estimated depth maps and camera parameters produces more accurate 3D points compared to directly employing a specialized point map branch.

\subsection{Feature Backbone}%
\label{sec:architecture}

Following recent works in 3D deep learning~\cite{wei24meshlrm:,wang24dust3r:,jin24lvsm:}, we design a simple architecture with minimal 3D inductive biases, letting the model learn from ample quantities of 3D-annotated data.
In particular, we implement the model $f$ as a large transformer~\cite{vaswani17attention}.
To this end, each input image $I$ is initially patchified into a set of $K$ tokens\footnote{The number of tokens depends on the image resolution.} $\tok^I \in \R^{K\times C}$ through DINO~\cite{oquab24dinov2:}. 
%
%
The combined set of image tokens from all frames, \ie, $\tok^I = \cup_{i=1}^N \{\tok^I_i\}$, is subsequently processed through the main network structure, alternating frame-wise and global self-attention layers.

\paragraph{Alternating-Attention.}

We slightly adjust the standard transformer design by introducing {Alternating-Attention} (AA), making the transformer focus within each frame and globally in an alternate fashion.
Specifically, \textit{frame-wise self-attention} attends to the tokens $\tok^I_k$ within each frame separately, and \textit{global self-attention} attends to the tokens $\tok^I$ across all frames jointly. 
This strikes a balance between integrating information across different images and normalizing the activations for the tokens within each image.
By default, we employ $L=24$ layers of global and frame-wise attention.
In \cref{sec:exp}, we demonstrate that our AA architecture brings significant performance gains.
Note that our architecture does not employ any cross-attention layers, only self-attention ones.

\subsection{Prediction heads}%
\label{sec:prediction_heads}

Here, we describe how $f$ predicts the camera parameters, depth maps, point maps, and point tracks.
First, for each input image $I_i$, we augment the corresponding image tokens $\tok^I_i$ with an additional camera token $\tok^\bg_i \in \R^{1 \times C'}$ and four register tokens~\cite{registers} $\tok^R_i \in \R^{4 \times C'}$.
The concatenation of $(\tok^I_i, \tok^\bg_i, \tok^R_ij)_{i=1}^{N}$ is then passed to the AA transformer, yielding output tokens $({\hat\tok}^I_i, {\hat\tok}^\bg_i, {\hat\tok}^R_i)_{i=1}^N$.
Here, the camera token and register tokens of the first frame ($\tok^\bg_1 := \bar{\tok}^\bg$, $\tok^R_1 := \bar{\tok}^R$) are set to a different set of learnable tokens $\overline{\tok}^\bg, \overline{\tok}^R$ than those of all other frames ($\tok^\bg_i := \overline{\overline{\tok}}^\bg, \tok^R_i := \overline{\overline{\tok}}^R, i \in [2,\dots,N]$), which are also learnable.
This allows the model to distinguish the first frame from the rest, and to represent the 3D predictions in the coordinate frame of the first camera.
Note that the refined camera and register tokens now become frame-specific—--this is because our AA transformer contains frame-wise self-attention layers that allow the transformer to match the camera and register tokens with the corresponding tokens from the same image.
Following common practice, the output register tokens ${\hat\tok}^R_i$ are discarded while ${\hat\tok}^I_i$, ${\hat\tok}^\bg_i$ are used for prediction.

\paragraph{Coordinate Frame.}
As noted above, we predict cameras, point maps, and depth maps in the coordinate frame of the first camera $\bg_1$.
As such, the camera extrinsics output for the first camera are set to the identity, \ie, the first rotation quaternion is $\bq_1 = [0, 0, 0, 1]$ and the first translation vector is $\bt_1 = [0, 0, 0]$.
Recall that the special camera and register tokens $\tok^\bg_1 := \overline{\tok}^\bg, \tok^R_1 := \overline{\tok}^R$ allow the transformer to identify the first camera.

\paragraph{Camera Predictions.}
The camera parameters $(\hat{\bg}^i)_{i=1}^N$ are predicted from the output camera tokens $({\hat\tok}^\bg_i)_{i=1}^N$ using four additional self-attention layers followed by a linear layer.
This forms the \emph{camera head} that predicts the camera intrinsics and extrinsics.

\paragraph{Dense Predictions.} 
The output image tokens ${\hat\tok}^I_i$ are used to predict the dense outputs, \ie, the depth maps $D_i$, point maps $P_i$, and tracking features $T_i$.
More specifically, ${\hat\tok}^I_i$ are first converted to dense feature maps $F_i \in \R^{C'' \times H \times W}$ with a DPT layer~\cite{ranftl21dpt}.
Each $F_i$ is then mapped with a $3\times 3$ convolutional layer to the corresponding depth and point maps $D_i$ and $P_i$.
Additionally, the DPT head also outputs dense features $T_i \in \R^{C \times H \times W}$, which serve as input to the tracking head.
We also predict the aleatoric uncertainty~\cite{kendall16modelling,novotny18capturing}
$\Sigma_i^D \in \R_+^{H \times W}$ and
$\Sigma_i^P \in \R_+^{H \times W}$ for each depth and point map, respectively.
As described in \cref{sec:training_losses}, the uncertainty maps are used in the loss and, after training, are proportional to the model's confidence in the predictions.

\paragraph{Tracking.}
In order to implement the tracking module $\mathcal{T}$, we use the CoTracker2 architecture~\cite{karaev24cotracker}, which takes the dense tracking features $T_i$ as input.
More specifically, given a query point $\by_j$ in a query image $I_q$ (during training, we always set $q=1$, but any other image can be potentially used as a query), the tracking head $\mathcal{T}$ predicts the set of 2D points
$
\mathcal{T}((\mathbf{y}_j)_{j=1}^M, (T_i)_{i=1}^N)
= ((\hat{\by}_{j,i})_{i=1}^N)_{j=1}^M
$
in all images $I_i$ that correspond to the same 3D point as $\by$.
To do so, the feature map $T_q$ of the query image is first bilinearly sampled at the query point $\by_j$ to obtain its feature.
This feature is then correlated with all other feature maps $T_i, i \neq q$ to obtain a set of correlation maps.
These maps are then processed by self-attention layers to predict the final 2D points $\hat{\by}_i$, which are all in correspondence with $\by_j$.
Note that, similar to VGGSfM~\cite{wang24vggsfm:}, our tracker does not assume any temporal ordering of the input frames and, hence, can be applied to any set of input images, not just videos.

\subsection{Training}\label{sec:training_losses}

\paragraph{Training Losses.}
We train the \themethod model $f$ end-to-end using a multi-task loss:
\begin{equation}\label{eq:training_loss}
\mathcal{L}
=
\mathcal{L}_\text{camera}
+ \mathcal{L}_\text{depth}
+ \mathcal{L}_\text{pmap}
+ \lambda \mathcal{L}_\text{track}.
\end{equation}
We found that the camera ($\mathcal{L}_\text{camera}$), depth ($\mathcal{L}_\text{depth}$), and point-map ($\mathcal{L}_\text{pmap}$) losses have similar ranges and do not need to be weighted against each other.
The tracking loss $\mathcal{L}_\text{track}$ is down-weighted with a factor of $\lambda = 0.05$.
We describe each loss term in turn.

The camera loss $\mathcal{L}_\text{camera}$ supervises the cameras $\hat{\bg}$:
$
\mathcal{L}_\text{camera} = \sum_{i=1}^N 
\left \| \hat{\bg}_i - \bg_i 
\right \|_\epsilon,
$
comparing the predicted cameras $\hat{\bg}_i$ with the ground truth $\bg_i$ using the Huber loss $|\cdot|_\epsilon$.

The depth loss $\mathcal{L}_\text{depth}$ follows \duster~\cite{wang24dust3r:} and implements the aleatoric-uncertainty loss~\cite{novotny17learning,kendall17what} weighing the discrepancy between the predicted depth $\hat{D}_i$ and the ground-truth depth $D_i$ with the predicted uncertainty map $\hat{\Sigma}_i^D$.
Differently from \duster, we also apply a gradient-based term, which is widely used in monocular depth estimation.
Hence, the depth loss is
$
\mathcal{L}_\text{depth}
=
\sum_{i=1}^N
\| \Sigma_i^D \odot (\hat{D}_i - D_i) \| + \| \Sigma_i^D \odot ({\smallnabla} \hat{D}_i - {\smallnabla} D_i) \|
- \alpha \log \Sigma_i^D
$,
where $\odot$ is the channel-broadcast element-wise product. 
The point map loss is defined analogously but with the point-map uncertainty $\Sigma_i^P$:
$
\mathcal{L}_\text{pmap}
=
\sum_{i=1}^N
\| \Sigma_i^P \odot (\hat{P}_i - P_i) \| + \| \Sigma_i^P \odot ({\smallnabla} \hat{P}_i - {\smallnabla}  P_i) \|
- \alpha \log \Sigma_i^P$.

Finally, the tracking loss is given by
$
\mathcal{L}_\text{track}
=
\sum_{j=1}^M
\sum_{i=1}^N
\|
    \by_{j,i}
    -
    \hat{\by}_{j,i}
\|
$.
Here, the outer sum runs over all ground-truth query points $\by_j$ in the query image $I_q$, $\by_{j,i}$ is $\by_j$'s ground-truth correspondence in image $I_i$, and $\hat{\by}_{j,i}$ is the corresponding prediction obtained by the application $\mathcal{T}((\mathbf{y}_j)_{j=1}^M, (T_i)_{i=1}^N)$ of the tracking module.
Additionally, following CoTracker2~\cite{karaev24cotracker}, we apply a visibility loss (binary cross-entropy) to estimate whether a point is visible in a given frame. 

\paragraph{Ground Truth Coordinate Normalization.}
If we scale a scene or change its global reference frame, the images of the scene are not affected at all, meaning that any such variant is a legitimate result of 3D reconstruction.
We remove this ambiguity by normalizing the data, thus making a canonical choice and task the transformer to output this particular variant.
We follow~\cite{wang24dust3r:} and, first, express all quantities in the coordinate frame of the first camera $\bg_1$.
Then, we compute the average Euclidean distance of all 3D points in the point map $P$ to the origin and use this scale to normalize the camera translations $\bt$, the point map $P$, and the depth map $D$.
Importantly, unlike~\cite{wang24dust3r:}, we do \emph{not} apply such normalization to the predictions output by the transformer; instead, we force it to learn the normalization we choose from the training data.

\paragraph{Implementation Details.}
By default, we employ $L=24$ layers of global and frame-wise attention, respectively.
The model consists of approximately 1.2 billion parameters in total.
We train the model by optimizing the training loss~\eqref{eq:training_loss} with the AdamW optimizer for $160$K iterations.
We use a cosine learning rate scheduler with a peak learning rate of $0.0002$ and a warmup of $8$K iterations.
For every batch, we randomly sample $2$--$24$ frames from a random training scene.
The input frames, depth maps, and point maps are resized to a maximum dimension of $518$ pixels.
The aspect ratio is randomized between 0.33 and 1.0.
We also randomly apply color jittering, Gaussian blur, and grayscale augmentation to the frames.
The training runs on 64 A100 GPUs over nine days.
We employ gradient norm clipping with a threshold of $1.0$ to ensure training stability.
We leverage bfloat16 precision and gradient checkpointing to improve GPU memory and computational efficiency.

\paragraph{Training Data.}
The model was trained using a large and diverse collection of datasets, including:
Co3Dv2~\cite{reizenstein21common},
BlendMVS~\cite{yao2020blendedmvs},
DL3DV~\cite{ling2024dl3dv},
MegaDepth~\cite{li2018megadepth},
Kubric~\cite{greff22kubric:},
WildRGB~\cite{xia2024rgbd},
ScanNet~\cite{dai2017scannet},
HyperSim~\cite{hypersim},
Mapillary~\cite{MPSD_2020_ECCV},
Habitat~\cite{szot2021habitat},
Replica~\cite{straub2019replica},
MVS-Synth~\cite{mvssynth},
PointOdyssey~\cite{zheng2023point},
Virtual KITTI~\cite{cabon2020virtual},
Aria Synthetic Environments~\cite{Pan_2023_ICCV},
Aria Digital Twin~\cite{Pan_2023_ICCV},
and a synthetic dataset of artist-created assets similar to Objaverse~\cite{deitke2023objaverse}.
These datasets span various domains, including indoor and outdoor environments, and encompass synthetic and real-world scenarios.
The 3D annotations for these datasets are derived from multiple sources, such as direct sensor capture, synthetic engines, or SfM techniques~\cite{schonberger16structure-from-motion}.
The combination of our datasets is broadly comparable to those of  MASt3R~\cite{duisterhof24mast3r-sfm:} in size and diversity.

\section{Experiments}%
\label{sec:exp}

\begin{table}[t]
    \centering
    \footnotesize
    \resizebox{\columnwidth}{!}{%
    \begin{tabular}{c|c|c|c}
    \toprule
    \multirow{2}{*}{Methods} & \multicolumn{1}{c|}{Re10K \textit{(unseen)} } & \multicolumn{1}{c|}{CO3Dv2} & \multirow{2}{*}{Time}\\
                                                & AUC@30 $\uparrow$ & AUC@30 $\uparrow$ \\ 
    \midrule
    Colmap+SPSG~\cite{sarlin2020superglue}    & 45.2              & 25.3              &  $\sim$ 15s          \\
    PixSfM~\cite{lindenberger21pixel-perfect} & 49.4              & 30.1              &  $>$ 20s             \\
    PoseDiff~\cite{wang23posediffusion:}      & 48.0              & 66.5              &  $\sim$ 7s           \\
    DUSt3R~\cite{wang24dust3r:}               & 67.7              & 76.7              &  $\sim$ 7s           \\
    MASt3R~\cite{mast3r}                      & 76.4              & 81.8              &  $\sim$ 9s           \\
    VGGSfM v2~\cite{wang24vggsfm:}            & 78.9              & 83.4              &  $\sim$ 10s          \\ 
    \midrule
    MV-DUSt3R~\cite{tang2024mv} $^{\mathbf{\ddagger}}$ & 71.3 & 69.5 & {$\sim$ 0.6s} \\
    CUT3R~\cite{cut3r} $^{\mathbf{\ddagger}}$ & 75.3 & 82.8 & {$\sim$ 0.6s}  \\
    FLARE~\cite{zhang2025flare} $^{\mathbf{\ddagger}}$ & 78.8 & 83.3 & \underline{{$\sim$ 0.5s}}  \\
    Fast3R~\cite{yang2025fast3r} $^{\mathbf{\ddagger}}$  &72.7 & 82.5 & \textbf{$\sim$ 0.2s}  \\
    \midrule
    Ours (Feed-Forward)                       & \underline{85.3}  & \underline{88.2}  & \textbf{$\sim$ 0.2s} \\
    Ours (with BA)                            & \textbf{93.5}     & \textbf{91.8}     &  $\sim$ 1.8s         \\
    \bottomrule
    \end{tabular}
    }%
    \caption{%
    \textbf{Camera Pose Estimation on RealEstate10K~\cite{zhou2018stereo} and CO3Dv2~\cite{reizenstein21common}}
    with 10 random frames.
    All metrics the higher the better. 
    None of the methods were trained on the Re10K dataset. 
    Runtime were measured using one H100 GPU\@. 
    Methods marked with $^{\mathbf{\ddagger}}$ represent concurrent work.
    }\label{tab:pose}%
    \normalsize
    \end{table}
    
\begin{table}
\centering
\small
\newcommand{\tickmark}{\ding{51}}
\newcommand{\xmark}{\ding{55}}
\footnotesize 
\begin{tabular}{ccccc}
\toprule
    Known GT
    & \multirow{2}{*}{Method}
    & \multirow{2}{*}{Acc.$\downarrow$}
    & \multirow{2}{*}{Comp.$\downarrow$}
    & \multirow{2}{*}{Overall$\downarrow$}
    \\ 
    camera 
    &&&&\\\midrule
    \tickmark  &   Gipuma~\cite{gipuma} &     \textbf{0.283} &       0.873 &    0.578 \\
    \tickmark &       MVSNet~\cite{mvsnet} &     0.396 &       0.527 &    0.462 \\
    \tickmark&       CIDER~\cite{cider} &     0.417 &       0.437 &    0.427 \\
    \tickmark& PatchmatchNet~\cite{pathcmatchnet} &     0.427 &       0.377 &    0.417 \\
    \tickmark&  MASt3R~\cite{mast3r} &     0.403 &       0.344 &    0.374 \\
    \tickmark&  GeoMVSNet~\cite{geomvsnet} &     0.331 &\textbf{0.259} &    \textbf{0.295} \\ \midrule
  \xmark&     DUSt3R~\cite{wang24dust3r:} &     2.677 &       0.805 &    1.741 \\
\xmark & Ours & \textbf{0.389} & \textbf{0.374} & \textbf{0.382} \\
\bottomrule
\end{tabular}
\caption{
\textbf{Dense MVS Estimation on the DTU~\cite{dtudataset} Dataset.} 
Methods operating with known ground-truth camera are in the top part of the table, while the bottom part contains the methods that do not know the ground-truth camera.
}\label{tab:dtu}
\end{table}

\begin{table}
\centering
\footnotesize
\begin{tabular}{cccccc}
\toprule
Methods            & Acc.$\downarrow$  & Comp.$\downarrow$ & Overall$\downarrow$ &  Time       \\
\midrule
DUSt3R             & 1.167             & 0.842             & 1.005               &  $\sim$ 7s  \\
MASt3R             & 0.968             & 0.684             & 0.826               &  $\sim$ 9s  \\
Ours (Point)       & \underline{0.901} & \underline{0.518} & \underline{0.709}   & $\sim$ 0.2s \\
Ours (Depth + Cam) &\textbf{0.873}     & \textbf{0.482}    & \textbf{0.677}      &  $\sim$ 0.2s\\
\bottomrule
\end{tabular}
\caption{
\textbf{Point Map Estimation on ETH3D~\cite{eth3d}.}
DUSt3R and MASt3R use global alignment while ours is feed-forward and, hence, much faster. The row \textit{Ours (Point)} indicates the results using the point map head directly, while \textit{Ours (Depth + Cam)} denotes constructing point clouds from the depth map head combined with the camera head.}%
\label{tab:eth3d}
\end{table}

\begin{table}
\footnotesize
\centering
\begin{tabular}{lcccc}
\toprule
Method                                               & AUC@5 $\uparrow$ & AUC@10 $\uparrow$ & AUC@20 $\uparrow$ \\
\midrule
SuperGlue~\citep{sarlin2020superglue}                & 16.2             & 33.8              & 51.8              \\
    LoFTR~\citep{sun2021loftr}                       & 22.1             & 40.8              & 57.6              \\
    DKM~\citep{edstedt2023dkm}                       & 29.4             & 50.7              & 68.3              \\
CasMTR~\citep{cao2023casmtr}                         & 27.1             & 47.0              & 64.4              \\        
    Roma~\citep{edstedt2024roma}                     & {31.8}           & {53.4}            & {70.9}            \\ \midrule
    Ours                                             & \textbf{33.9}    & \textbf{55.2}     & \textbf{73.4}     \\
\bottomrule
\end{tabular}
\caption{\textbf{Two-View matching comparison on ScanNet-1500~\citep{dai2017scannet, sarlin2020superglue}}. Although our tracking head is not specialized for the two-view setting, it outperforms the state-of-the-art two-view matching method Roma. Measured in AUC (higher is better).}%
\label{tab:scannet}
\end{table}

This section compares our method to state-of-the-art approaches across multiple tasks to show its effectiveness.

\subsection{Camera Pose Estimation}

We first evaluate our method on the CO3Dv2~\cite{reizenstein21common} and RealEstate10K~\cite{zhou2018stereo} datasets for camera pose estimation, as shown in~\cref{tab:pose}.
Following~\cite{wang23posediffusion:}, we randomly select 10 images per scene and evaluate them using the standard metric AUC@30, which combines RRA and RTA\@.
RRA (Relative Rotation Accuracy) and RTA (Relative Translation Accuracy) calculate the relative angular errors in rotation and translation, respectively, for each image pair.
These angular errors are then thresholded to determine the accuracy scores.
AUC is the area under the accuracy-threshold curve of the minimum values between RRA and RTA across varying thresholds.
The (learnable) methods in~\cref{tab:pose} have been trained on Co3Dv2 and \textbf{not} on RealEstate10K.
Our feed-forward model consistently outperforms competing methods across all metrics on both datasets, including those that employ computationally expensive post-optimization steps, such as Global Alignment for \duster/\master and Bundle Adjustment for VGGSfM, typically requiring more than $10$ seconds.
In contrast, \themethod \emph{achieves superior performance while only operating in a feed-forward manner}, requiring just $0.2$ seconds on the same hardware.
Compared to concurrent works~\cite{yang2025fast3r, tang2024mv, zhang2025flare, cut3r} (indicated by $^{\mathbf{\ddagger}}$), our method demonstrates significant performance advantages, with speed similar to the fastest variant Fast3R~\cite{yang2025fast3r}.
Furthermore, our model's performance advantage is even more pronounced on the RealEstate10K dataset, which none of the methods presented in \cref{tab:pose} were trained on.
This validates the superior generalization of \themethod.

Our results also show that \themethod can be improved even further by combining it with optimization methods from visual geometry optimization like BA\@.
Specifically, refining the predicted camera poses and tracks with BA further improves accuracy.
Note that our method directly predicts close-to-accurate point/depth maps, which can serve as a good initialization for BA\@.
This eliminates the need for triangulation and iterative refinement in BA as done by~\cite{wang24vggsfm:}, making our approach significantly faster (only around $2$ seconds even with BA).
Hence, while the feed-forward mode of \themethod outperforms all previous alternatives (whether they are feed-forward or not), there is still room for improvement since post-optimization still brings benefits.

\begin{figure*}[htbp] 
\includegraphics[width=1.0\linewidth]{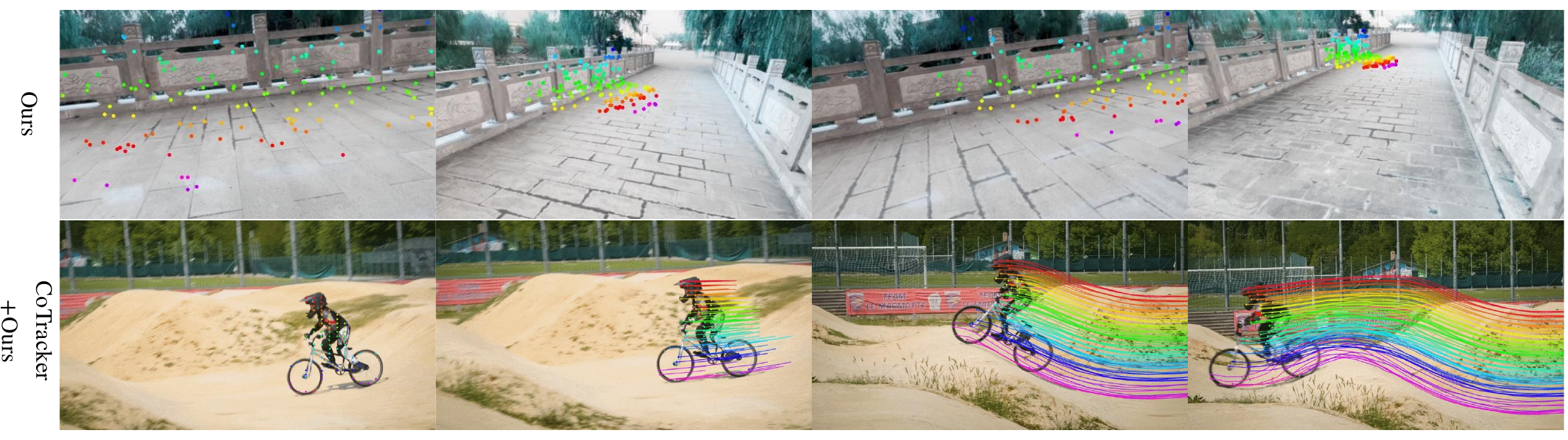}
\caption{\textbf{Visualization of Rigid and Dynamic Point Tracking.}  
Top: \themethod's tracking module $\mathcal{T}$ outputs keypoint tracks for an unordered set of input images depicting a static scene. 
Bottom: We finetune the backbone of \themethod to enhance a dynamic point tracker  CoTracker~\cite{karaev2023cotracker}, which processes sequential inputs.  
}%
\label{fig:dynamic_track}
\end{figure*}

\subsection{Multi-view Depth Estimation}

Following \master~\cite{mast3r}, we further evaluate our multi-view depth estimation results on the DTU~\cite{dtudataset} dataset.
We report the standard DTU metrics, including Accuracy (the smallest Euclidean distance from the prediction to ground truth), Completeness (the smallest Euclidean distance from the ground truth to prediction), and their average Overall (\ie, Chamfer distance).
In \cref{tab:dtu}, \duster and our \themethod are the only two methods operating without the knowledge of ground truth cameras.
\master derives depth maps by triangulating matches using the ground truth cameras.
Meanwhile, deep multi-view stereo methods like GeoMVSNet use ground truth cameras to construct cost volumes.

Our method substantially outperforms \duster, reducing the Overall score from $1.741$ to $0.382$.
More importantly, it achieves results comparable to methods that know ground-truth cameras at test time.
The significant performance gains can likely be attributed to our model's multi-image training scheme that teaches it to reason about multi-view triangulation natively, instead of relying on ad hoc alignment procedures, such as in \duster, which only averages multiple pairwise camera triangulations.

\subsection{Point Map Estimation}

We also compare the accuracy of our predicted point cloud to \duster and \master on the ETH3D~\cite{eth3d} dataset.
For each scene, we randomly sample $10$ frames.
The predicted point cloud is aligned to the ground truth using the Umeyama~\cite{umeyama91least-squares} algorithm.
The results are reported after filtering out invalid points using the official masks.
We report Accuracy, Completeness, and Overall (Chamfer distance) for point map estimation.
As shown in \cref{tab:eth3d}, although \duster and \master conduct expensive optimization (global alignment--—around 10 seconds per scene), our method still outperforms them significantly in a simple feed-forward regime at only $0.2$ seconds per reconstruction.

Meanwhile, compared to directly using our estimated point maps, we found that the predictions from our depth and camera heads (\ie, unprojecting the predicted depth maps to 3D using the predicted camera parameters) yield higher accuracy.
We attribute this to the benefits of decomposing a complex task (point map estimation) into simpler subproblems (depth map and camera prediction), even though camera, depth maps, and point maps are jointly supervised during training.

We present a qualitative comparison with \duster on in-the-wild scenes in \cref{fig:pointmap_compare} and further examples in \cref{fig:pointmap}.
\themethod outputs high-quality predictions and generalizes well, excelling on challenging out-of-domain examples, such as oil paintings, non-overlapping frames, and scenes with repeating or homogeneous textures like deserts.

\subsection{Image Matching}

Two-view image matching is a widely-explored topic~\cite{sarlin20superglue:,lindenberger23lightglue:, sun2021loftr} in computer vision.
It represents a specific case of rigid point tracking, which is restricted to only two views, and hence a suitable evaluation benchmark to measure our tracking accuracy, even though our model is not specialized for this task.
We follow the standard protocol~\cite{edstedt2024roma,sarlin20superglue:} on the ScanNet dataset~\cite{dai2017scannet} and report the results in \cref{tab:scannet}.
For each image pair, we extract the matches and use them to estimate an essential matrix, which is then decomposed to a relative camera pose.
The final metric is the relative pose accuracy, measured by AUC\@.
For evaluation, we use ALIKED~\cite{zhao2023aliked} to detect keypoints, treating them as query points $\by_q$.
These are then passed to our tracking branch $\mathcal{T}$ to find correspondences in the second frame.
We adopt the evaluation hyperparameters (\eg, the number of matches, RANSAC thresholds) from Roma~\cite{edstedt2024roma}.
Despite not being explicitly trained for two-view matching, \cref{tab:scannet} shows that \themethod achieves the highest accuracy among all baselines.

\begin{table} 
\centering
\footnotesize 
\begin{tabular}{c|ccc}
\toprule
ETH3D Dataset              & Acc.$\downarrow$  & Comp.$\downarrow$ & Overall$\downarrow$ \\
\midrule
Cross-Attention            & 1.287             & 0.835             & 1.061               \\
Global Self-Attention Only & \underline{1.032 }& \underline{0.621} & \underline{0.827}   \\
Alternating-Attention      & \textbf{0.901}    & \textbf{0.518}    & \textbf{0.709}      \\
\bottomrule
\end{tabular}
\caption{
\textbf{Ablation Study for Transformer Backbone} on ETH3D. We compare our alternating-attention architecture against two variants: one using only global self-attention and another employing cross-attention. 
}%
\label{tab:ablation}
\vspace{-6pt}
\end{table}

\begin{table} 
\centering
\footnotesize
\resizebox{0.9\columnwidth}{!}{
\newcommand{\tickmark}{\ding{51}}
\newcommand{\xmark}{\ding{55}} 
\begin{tabular}{ccccccc}
\toprule
w. $\mathcal{L}_\text{camera}$   & w. $\mathcal{L}_\text{depth}$  & w. $\mathcal{L}_\text{track}$       & Acc.$\downarrow$  &  Comp.$\downarrow$ &  Overall$\downarrow$ \\
\midrule
\xmark      & \tickmark & \tickmark      & 1.042             &  0.627             &  0.834               \\
 \tickmark  & \xmark    & \tickmark      & \underline{0.920} &  \underline{0.534} &  \underline{0.727}   \\ 
 \tickmark  & \tickmark & \xmark         & 0.976             &  0.603             &  0.790               \\
\tickmark   & \tickmark & \tickmark      & \textbf{0.901}    &  \textbf{0.518}    &  \textbf{0.709}      \\
\bottomrule
\end{tabular}
}
\caption{
\textbf{Ablation Study for Multi-task Learning}, which shows that simultaneous training with camera, depth and track estimation yields the highest accuracy in point map estimation on ETH3D. 
}%
\label{tab:multitask}
\end{table}

\subsection{Ablation Studies}%
\label{sec:ablation}

\paragraph{Feature Backbone.}

We first validate the effectiveness of our proposed Alternating-Attention design by comparing it against two alternative attention architectures: (a) \textit{global self-attention only}, and (b) \textit{cross-attention}.
To ensure a fair comparison, all model variants maintain an identical number of parameters, using a total of $2L$ attention layers.
For the cross-attention variant, each frame independently attends to tokens from all other frames, maximizing cross-frame information fusion although significantly increasing the runtime, particularly as the number of input frames grows.
The hyperparameters such as the hidden dimension and the number of heads are kept the same.
Point map estimation accuracy is chosen as the evaluation metric for our ablation study, as it reflects the model's joint understanding of scene geometry and camera parameters.
Results in \cref{tab:ablation} demonstrate that our Alternating-Attention architecture outperforms both baseline variants by a clear margin.
Additionally, our other preliminary exploratory experiments consistently showed that architectures using cross-attention generally underperform compared to those exclusively employing self-attention.

\paragraph{Multi-task Learning.}

We also verify the benefit of training a single network to simultaneously learn multiple 3D quantities, even though these outputs may potentially overlap (\eg, depth maps and camera parameters together can produce point maps).
As shown in \cref{tab:multitask}, there is a noticeable decrease in the accuracy of point map estimation when training without camera, depth, or track estimation.
Notably, incorporating camera parameter estimation clearly enhances point map accuracy, whereas depth estimation contributes only marginal improvements.

\subsection{Finetuning for Downstream Tasks}

We now show that the \themethod pre-trained feature extractor can be reused in downstream tasks.
We show this for feed-forward novel view synthesis and dynamic point tracking.

\begin{figure}[htbp]
\includegraphics[width=1.0\linewidth]{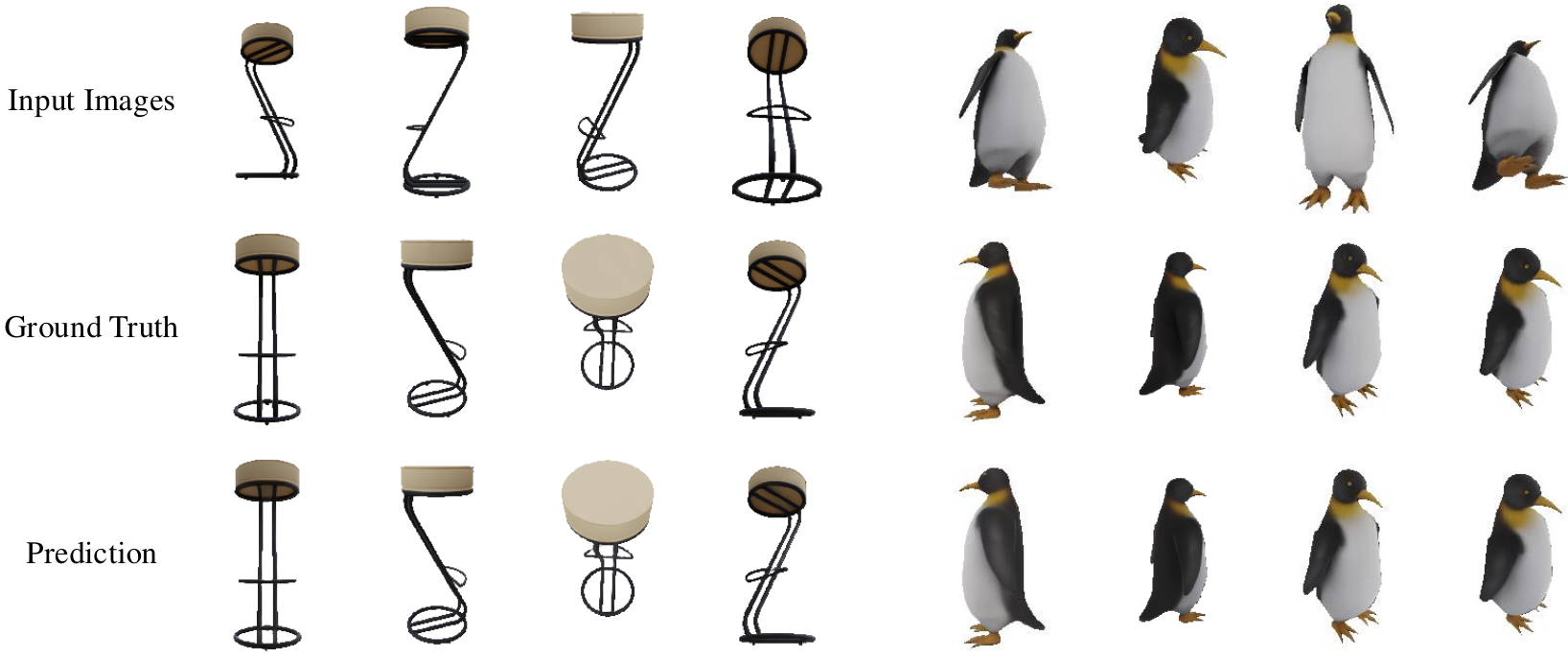}
\caption{\textbf{Qualitative Examples of Novel View Synthesis.} The top row shows the input images, the middle row displays the ground truth images from target viewpoints, and the bottom row presents our synthesized images.}
\label{fig:nvs}
\end{figure}

\begin{table}[htbp]
\footnotesize
\newcommand{\tickmark}{\ding{51}}
\newcommand{\xmark}{\ding{55}}
\begin{center}
\resizebox{0.95\columnwidth}{!}{%
\begin{tabular}{ccc|ccc} 
\midrule
Method                     & \scriptsize{Known Input Cam} & Size & PSNR \(\uparrow \) & SSIM \(\uparrow \) & LPIPS \(\downarrow \) \\ 
\midrule
LGM~\citep{tang2024lgm} 
                           & \tickmark                    & 256  & 21.44              & 0.832              & 0.122                 \\
GS-LRM~\citep{zhang2024gs} & \tickmark                    & 256  & 29.59              & 0.944              & 0.051                 \\
LVSM~\cite{jin24lvsm:}     & \tickmark                    & 256  & 31.71              & 0.957              & 0.027                 \\
\midrule
{Ours-NVS}$^*$             & \xmark                       & 224  & 30.41             & 0.949              & 0.033      
\\
\midrule
\end{tabular}}
\caption{\textbf{Quantitative comparisons for view synthesis on GSO~\cite{downs2022google} dataset.} 
Finetuning \themethod for feed-forward novel view synthesis, it demonstrates competitive performance even without knowing camera extrinsic and intrinsic parameters for the input images. 
Note that $^*$ indicates using a small training set (only 20\%).}%
\label{tab:compare_gso}
\end{center}
\vspace{-4pt}
\end{table}

\paragraph{Feed-forward Novel View Synthesis} is progressing rapidly~\cite{hong24lrm:, wang23pf-lrm:, cao25lightplane:, xu24grm:, zhang24gs-lrm:, szymanowicz2024splatter, han2024flex3d, jin24lvsm:}. Most existing methods take images with known camera parameters as input and predict the target image corresponding to a new camera viewpoint.
Instead of relying on an explicit 3D representation, we follow LVSM~\cite{jin24lvsm:} and modify \themethod to \emph{directly} output the target image.
However, we \emph{do not} assume known camera parameters for the input frames.

We follow the training and evaluation protocol of LVSM closely, \eg, using $4$ input views and adopting Plücker rays to represent target viewpoints.
We make a simple modification to \themethod.
As before, the input images are converted into tokens by DINO\@.
Then, for the target views, we use a convolutional layer to encode their Plücker ray images into tokens.
These tokens, representing both the input images and the target views, are concatenated and processed by the AA transformer.
Subsequently, a DPT head is used to regress the RGB colors for the target views.
It is important to note that we do \emph{not} input the Plücker rays for the source images.
Hence, the model is not given the camera parameters for these input frames.

LVSM was trained on the Objaverse dataset~\cite{deitke2023objaverse}.
We use a similar internal dataset of approximately 20\% the size of Objaverse.
Further details on training and evaluation can be found in~\cite{jin24lvsm:}.
As shown in \cref{tab:compare_gso}, despite not requiring the input camera parameters and using less training data than LVSM, our model achieves competitive results on the GSO dataset~\cite{downs2022google}.
We expect that better results would be obtained using a larger training dataset.
Qualitative examples are shown in \cref{fig:nvs}.


\begin{table}[t]
\centering
\setlength{\tabcolsep}{2pt}
\footnotesize
\begin{tabular}{lcccccccccccccccc}
\toprule
\multirow{2}{*}[-0.2em]{Method} & \multicolumn{3}{c}{Kinetics } & \multicolumn{3}{c}{RGB-S}  & \multicolumn{3}{c}{DAVIS}\\
\cmidrule(lr){2-4}
\cmidrule(lr){5-7}
\cmidrule(lr){8-10}
                                       & AJ               & \davgvis         & OA              & AJ               & \davgvis         & OA               & AJ               & \davgvis         & OA               \\ \midrule
TAPTR~\citep{li2024taptr}              & 49.0             & 64.4             & 85.2            & 60.8             & 76.2             & 87.0             & \underline{63.0} & \underline{76.1} & \underline{91.1} \\
LocoTrack~\citep{cho2024local}         & 52.9             & 66.8             & 85.3            & 69.7             & \underline{83.2} & \underline{89.5} & 62.9             & 75.3             & 87.2             \\
BootsTAPIR~\citep{doersch2024bootstap} & \underline{54.6} & \underline{68.4} &\underline{86.5} & \underline{70.8} & 83.0             & 89.9             & 61.4             &73.6              & 88.7             \\ \midrule
CoTracker~\citep{karaev2023cotracker}  & 49.6             & 64.3             & 83.3            & 67.4             &78.9              & 85.2             & 61.8             & 76.1             & 88.3             \\
CoTracker + Ours                       & \textbf{57.2}    & \textbf{69.0}    & \textbf{88.9}   & \textbf{72.1}    & \textbf{84.0}    & \textbf{91.6}    & \textbf{64.7}    & \textbf{77.5}    & \textbf{91.4}    \\
\bottomrule
\end{tabular}
\caption{\textbf{Dynamic Point Tracking Results on the TAP-Vid benchmarks.}
Although our model was not designed for dynamic scenes, simply fine-tuning CoTracker with our pretrained weights significantly enhances performance, demonstrating the robustness and effectiveness of our learned features. 
}%
\label{tab:tab_tapvid}
\end{table}

\paragraph{Dynamic Point Tracking}  has emerged as a highly competitive task in recent years~\cite{pips, doersch23tapir:, xiao2024spatialtracker, karaev24cotracker}, and it serves as another downstream application for our learned features.
Following standard practices, we report these point-tracking metrics:
Occlusion Accuracy (OA), which comprises the binary accuracy of occlusion predictions;
\davgvis, comprising the mean proportion of visible points accurately tracked within a certain pixel threshold; and
Average Jaccard (AJ), measuring tracking and occlusion prediction accuracy together.

We adapt the state-of-the-art CoTracker2 model~\cite{karaev24cotracker} by substituting its backbone with our pretrained feature backbone.
This is necessary because \themethod is trained on unordered image collections instead of sequential videos.
Our backbone predicts the tracking features $T_i$, which replace the outputs of the feature extractor and later enter the rest of the CoTracker2 architecture, that finally predicts the tracks.
We finetune the entire modified tracker on Kubric~\cite{greff22kubric:}.
As illustrated in \cref{tab:tab_tapvid}, the integration of pretrained \themethod significantly enhances CoTracker's performance on the TAP-Vid benchmark~\cite{tapvid}.
For instance, \themethod's tracking features improve the \davgvis metric from $78.9$ to $84.0$ on the TAP-Vid RGB-S dataset.
Despite the TAP-Vid benchmark's inclusion of videos featuring rapid dynamic motions from various data sources, our model's strong performance demonstrates the generalization capability of its features, even in scenarios for which it was not explicitly designed.

\section{Discussions}

\paragraph{Limitations.}

While our method exhibits strong generalization to diverse in-the-wild scenes, several limitations remain.
First, the current model does not support fisheye or panoramic images. Additionally, reconstruction performance drops under conditions involving extreme input rotations.
Moreover, although our model handles scenes with minor non-rigid motions, it fails in scenarios involving substantial non-rigid deformation.

However, an important advantage of our approach is its flexibility and ease of adaptation. Addressing these limitations can be straightforwardly achieved by fine-tuning the model on targeted datasets with minimal architectural modifications. This adaptability clearly distinguishes our method from existing approaches, which typically require extensive re-engineering during test-time optimization to accommodate such specialized scenarios.

\begin{table}[t]
\footnotesize
\begin{center}
\resizebox{\columnwidth}{!}{
\begin{tabular}{c|ccccccccc}
\hline
\textbf{Input Frames} & 1 & 2 & 4 & 8 & 10 & 20 & 50 & 100 & 200 \\
\hline
\textbf{Time (s)} & 0.04 & 0.05 & 0.07 & 0.11 & 0.14 & 0.31 & 1.04 & 3.12 & 8.75 \\
\textbf{Mem. (GB)} & 1.88 & 2.07 & 2.45 & 3.23 & 3.63 & 5.58 & 11.41 & 21.15 & 40.63 \\
\hline
\end{tabular}
}
\caption{\textbf{Runtime and peak GPU memory usage across different numbers of input frames.} Runtime is measured in seconds, and GPU memory usage is reported in gigabytes.}%
\label{tab:runtime_mem}
\end{center}
\vspace{-10pt}
\end{table}

\paragraph{Runtime and Memory.}
As shown in \cref{tab:runtime_mem}, we evaluate inference runtime and peak GPU memory usage of the feature backbone when processing varying numbers of input frames. Measurements are conducted using a single NVIDIA H100 GPU with flash attention v3~\cite{shah2024flashattention}.
Images have a resolution of $336\times518$.

We focus on the cost associated with the feature backbone since users may select different branch combinations depending on their specific requirements and available resources. The camera head is lightweight, typically accounting for approximately $5\%$ of the runtime and about $2\%$ of the GPU memory used by the feature backbone. A DPT head uses an average of $0.03$ seconds and $0.2$ GB GPU memory per frame.

When GPU memory is sufficient, multiple frames can be processed efficiently in a single forward pass.
At the same time, in our model, inter-frame relationships are handled only within the feature backbone, and the DPT heads make independent predictions per frame.
Therefore, users constrained by GPU resources may perform predictions frame by frame. We leave this trade-off to the user's discretion.

We recognize that a naive implementation of global self-attention can be highly memory-intensive with a large number of tokens. Savings or accelerations can be achieved by employing techniques used in large language model (LLM) deployments. For instance, Fast3R~\cite{yang2025fast3r} employs Tensor Parallelism to accelerate inference with multiple GPUs, which can be directly applied to our model.

\paragraph{Patchifying.}

As discussed in \cref{sec:architecture}, we have explored the method of patchifying images into tokens by utilizing either a $14 \times 14$ convolutional layer or a pretrained DINOv2 model. Empirical results indicate that the DINOv2 model provides better performance; moreover, it ensures much more stable training, particularly in the initial stages. The DINOv2 model is also less sensitive to variations in hyperparameters such as learning rate or momentum. Consequently, we have chosen DINOv2 as the default method for patchifying in our model.

\paragraph{Differentiable BA.}

We also explored the idea of using differentiable bundle adjustment as in VGGSfM~\cite{wang24vggsfm:}.
In small-scale preliminary experiments, differentiable BA demonstrated promising performance.
However, a bottleneck is its computational cost during training.
Enabling differentiable BA in PyTorch using Theseus~\cite{pineda2022theseus} typically makes each training step roughly $4$ times slower, which is expensive for large-scale training.
While customizing a framework to expedite training could be a potential solution, it falls outside the scope of this work.
Thus, we opted not to include differentiable BA in this work, but we recognize it as a promising direction for large-scale unsupervised training, as it can serve as an effective supervision signal in scenarios lacking explicit 3D annotations.

\paragraph{Single-view Reconstruction.}

Unlike systems like \duster and \master that have to duplicate an image to create a pair, our model architecture inherently supports the input of a single image. In this case, global attention simply transitions to frame-wise attention. Although our model was not explicitly trained for single-view reconstruction, it demonstrates surprisingly good results. Some examples can be found in \cref{fig:pointmap_compare} and \cref{fig:pointmap_supp}. We strongly encourage trying our demo for better visualization.

\paragraph{Normalizing Prediction.}

As discussed in \cref{sec:training_losses}, our approach normalizes the ground truth using the average Euclidean distance of the 3D points.
While some methods, such as \duster, also apply such normalization to network predictions, our findings suggest that it is neither necessary for convergence nor advantageous for final model performance.
Furthermore, it tends to introduce additional instability during the training phase.

\section{Conclusions}
We present \themethodfull (\themethod), a feed-forward neural network that can directly estimate all key 3D scene properties for hundreds of input views.
It achieves state-of-the-art results in multiple 3D tasks, including camera parameter estimation, multi-view depth estimation, dense point cloud reconstruction, and 3D point tracking. 
Our simple, neural-first approach departs from traditional visual geometry-based methods, which rely on optimization and post-processing to obtain accurate and task-specific results.
The simplicity and efficiency of our approach make it well-suited for real-time applications, which is another benefit over optimization-based approaches.

\appendix
\appendix

\newpage
\section*{Appendix}

\noindent In the Appendix, we provide the following: 
\begin{itemize}
    \item formal definitions of key terms in \cref{sec:definitions}.  
    \item comprehensive implementation details, including architecture and training hyperparameters in \cref{sec:impl}.  
    \item additional experiments and discussions in \cref{sec:addexp}.  
    \item qualitative examples of single-view reconstruction in \cref{sec:qual}.  
    \item an expanded review of related works in \cref{sec:addrel}. 
\end{itemize}

\section{Formal Definitions}\label{sec:definitions}

In this section, we provide additional formal definitions that further ground the method section.

The camera extrinsics are defined in relation to the \emph{world reference frame}, which we take to be the coordinate system of the first camera.
We thus introduce two functions.
The first function
$
\gamma(\bg, \bp) = \bp'
$
applies the rigid transformation encoded by $\bg$ to a point $\bp$ in the world reference frame to obtain the corresponding point $\bp'$ in the camera reference frame.
The second function
$
\pi(\bg, \bp) = \by
$
further applies perspective projection, mapping the 3D point $\bp$ to a 2D image point $\by$.
We also denote the depth of the point as observed from the camera $\bg$ by
$
\pi^\text{D}(\bg, \bp) = d \in \R^+
$.

We model the scene as a collection of regular surfaces $S_i \subset \mathbb{R}^3$.
We make this a function of the $i$-th input image as the scene can change over time~\cite{zhang24monst3r:}.
The depth at pixel location $\by \in \mathcal{I}(I_i)$ is defined as the minimum depth of any 3D point $\bp$ in the scene that projects to $\by$, \ie,
$
D_i(\by) = \min \{
    \pi^D(\bg_i, \bp) : \bp \in S_i ~\wedge~ \pi(\bg_i, \bp) = \by
\}.
$
The point at pixel location $\by$ is then given by $P_i(\by) = \gamma(\bg, \bp)$, where $\bp \in S_i$ is the 3D point that minimizes the expression above, \ie,
$
\bp \in S_i
~\wedge~ \pi(\bg_i, \bp) = \by
~\wedge~ \pi^D(\bg_i, \bp) = D_i(\by)
$.


\section{Implementation Details}\label{sec:impl} 

\paragraph{Architecture.} As mentioned in the main paper, \themethod consists of $24$ attention blocks, each block equipped with one frame-wise self-attention layer and one global self-attention layer. 
Following the ViT-L model used in DINOv2~\cite{oquab24dinov2:}, each attention layer is configured with a feature dimension of $1024$ and employs $16$ heads. 
We use the official implementation of the attention layer from PyTorch, \ie, \textit{torch.nn.MultiheadAttention}, with flash attention enabled.
To stabilize training, we also use QKNorm~\cite{henry2020query} and LayerScale~\cite{touvron2021going} for each attention layer. 
The value of LayerScale is initialized with $0.01$. 
For image tokenization, we use DINOv2~\cite{oquab24dinov2:} and add positional embedding. 
As in~\cite{depth_anything_v2}, we feed the tokens from the $4$-th, $11$-th, $17$-th, and $23$-rd block into DPT~\cite{ranftl21dpt} for upsampling.

\paragraph{Training.}
To form a training batch, we first choose a random training dataset (each dataset has a different yet approximately similar weight, as in~\cite{wang24dust3r:}), and from the dataset, we then sample a random scene (uniformly).
During the training phase, we select between $2$ and $24$ frames per scene while maintaining the constant total of $48$ frames within each batch. 
For training, we use the respective training sets of each dataset.
We exclude training sequences containing fewer than $24$ frames.
RGB frames, depth maps, and point maps are first isotropically resized, so the longer size has $518$ pixels.
Then, we crop the shorter dimension (around the principal point) to a size between $168$ and $518$ pixels while remaining a multiple of the $14$-pixel patch size.
It is worth mentioning that we apply aggressive color augmentation independently across each frame within the same scene, enhancing the model’s robustness to varying lighting conditions. 
We build ground truth tracks following~\cite{wang24vggsfm:,edstedt2024roma, sun2021loftr}, which unprojects depth maps to 3D, reprojects points to target frames, and retains correspondences where reprojected depths match target depth maps. 
Frames with low similarity to the query frame are excluded during batch sampling. 
In rare cases with no valid correspondences, the tracking loss is omitted. 

\begin{figure*}[t] 
\centering
\includegraphics[width=\linewidth]{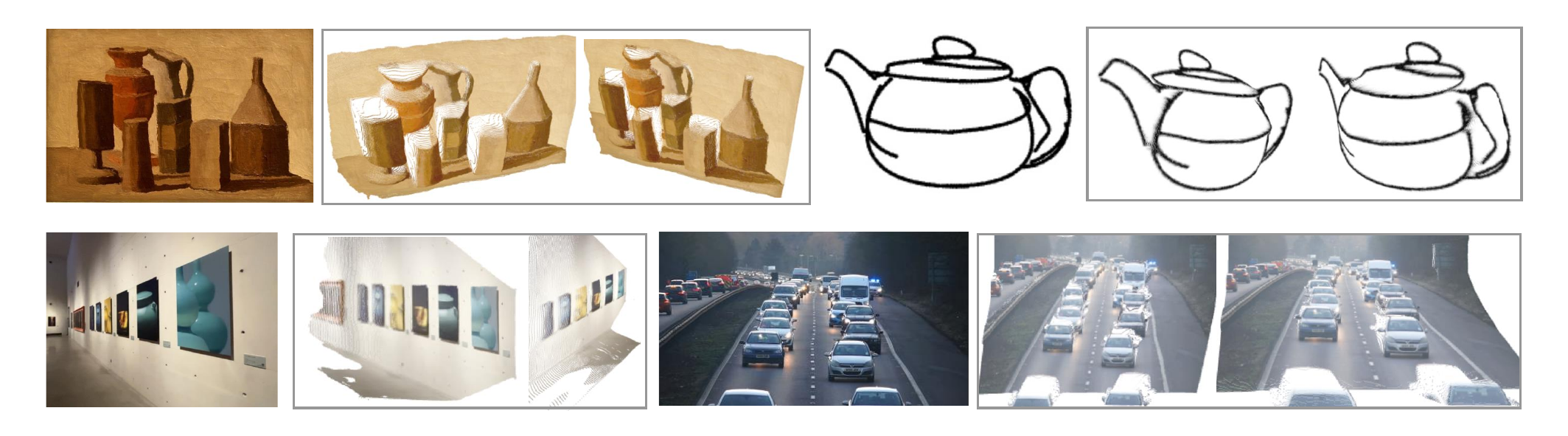}
\caption{\textbf{Single-view Reconstruction by Point Map Estimation.} Unlike DUSt3R, which requires duplicating an image into a pair, our model can predict the point map from a single input image. 
It demonstrates strong generalization to unseen real-world images.   
}%
\label{fig:pointmap_supp}
\end{figure*}

\section{Additional Experiments}\label{sec:addexp}

\paragraph{Camera Pose Estimation on IMC}
We also evaluate using the Image Matching Challenge (IMC)~\cite{jin2021image}, a camera pose estimation benchmark focusing on phototourism data. 
Until recently, the benchmark was dominated by classical incremental SfM methods~\cite{schoenberger2016sfm}.

\paragraph{Baselines.}
We evaluate two flavors of our model: \themethod and \themethod+ BA\@.
\themethod directly outputs camera pose estimates, while \themethod+ BA refines the estimates using an additional Bundle Adjustment stage.
We compare to the classical incremental SfM methods such as~\cite{schoenberger2016sfm,lindenberger21pixel-perfect} and to recently-proposed deep methods.
Specifically, recently VGGSfM~\cite{wang24vggsfm:} provided the first end-to-end trained deep method that outperformed incremental SfM on the challenging phototourism datasets.

Besides VGGSfM, we additionally compare to recently popularized DUSt3R~\cite{wang24dust3r:} and MASt3R~\cite{mast3r}.
It is important to note that DUSt3R and MASt3R utilized a substantial portion of the MegaDepth dataset for training, only excluding scenes $0015$ and $0022$. 
The MegaDepth scenes employed in their training have some overlap with the IMC benchmark, although the images are not identical; the same scenes are present in both datasets.
For instance, the MegaDepth scene $0024$ corresponds to the British Museum, while the British Museum is also a scene in the IMC benchmark.
For an apples-to-apples comparison, we adopt the same training split as DUSt3R and MASt3R. 
In the main paper, to ensure a fair comparison on ScanNet-1500, we exclude the corresponding ScanNet scenes from our training.

\paragraph{Results.}
\Cref{tab:IMCpose} contains the results of our evaluation.
Although phototourism data is the traditional focus of SfM methods, our \themethod's feed-forward performance is on par with the state-of-the-art VGGSfMv2 with AUC@10 of $71.26$ versus $76.82$, while being significantly faster (0.2 vs. 10 seconds per scene).
Remarkably, \themethod outperforms both MASt3R~\cite{mast3r} and DUSt3R~\cite{wang24dust3r:} significantly across all accuracy thresholds while being much faster.
This is because MASt3R's and DUSt3R's feed-forward predictions can only process pairs of frames and, hence, require a costly global alignment step.
Additionally, with bundle adjustment, \themethod+ BA further improves drastically, achieving state-of-the-art performance on IMC, raising AUC@10 from $71.26$ to $84.91$, and raising AUC@3 from $39.23$ to $66.37$.
Note that our model directly predicts 3D points, which can serve as the initialization for BA\@.
This eliminates the need for triangulation and iterative refinement of BA as in~\cite{wang24vggsfm:}.
As a result, \themethod+ BA is much faster than~\cite{wang24vggsfm:}.

\newcommand{\tickmark}{\ding{51}}
\newcommand{\xmark}{\ding{55}}
\begin{table}[tbp]
\centering
\small
\resizebox{\linewidth}{!}{
\begin{tabular}{cccccc} 
\toprule
Method                                                &  Test-time Opt. & AUC@3$\degree$    & AUC@5$\degree$    & AUC@10$\degree$   &
Runtime                                               \\ 
\midrule
COLMAP~(SIFT+NN)~\cite{schoenberger2016sfm}           &  \tickmark      & 23.58             & 32.66             & 44.79             & $>$10s        \\
PixSfM (SIFT + NN)~\cite{lindenberger21pixel-perfect} &  \tickmark      & 25.54             & 34.80             & 46.73             & $>$20s        \\
PixSfM (LoFTR)~\cite{lindenberger21pixel-perfect}     &  \tickmark      & 44.06             & 56.16             & 69.61             & $>$20s        \\
PixSfM (SP + SG)~\cite{lindenberger21pixel-perfect}   &  \tickmark      & {45.19}           & 57.22             & 70.47             & $>$20s        \\
DFSfM~(LoFTR)~\cite{he2023dfsfm}                      &  \tickmark      & {46.55}           & {58.74}           & {72.19}           & $>$10s        \\
\midrule
DUSt3R~\cite{wang24dust3r:}                           &  \tickmark      & 13.46             & 21.24             & 35.62             & $\sim$ 7s     \\
MASt3R~\cite{mast3r}                                  &  \tickmark      & 30.25             & 46.79             & 57.42             & $\sim$ 9s     \\
VGGSfM~\cite{wang24vggsfm:}                           &  \tickmark      & {45.23}           & {58.89}           & {73.92}           & $\sim$ 6s     \\
VGGSfMv2~\cite{wang24vggsfm:}                         &  \tickmark      & \underline{59.32} & \underline{67.78} & \underline{76.82} & $\sim$ 10s    \\
\midrule
\themethod (ours)                                     &  \xmark         & 39.23             & 52.74             & 71.26             & \textbf{0.2s} \\
\themethod + BA (ours)                                &  \tickmark      & \textbf{66.37}    & \textbf{75.16}    & \textbf{84.91}    & 1.8s          \\
\bottomrule
\end{tabular}
}
\caption{\textbf{Camera Pose Estimation on IMC~\cite{jin2021image}.} 
Our method achieves state-of-the-art performance on the challenging phototropism data, outperforming VGGSfMv2~\cite{wang24vggsfm:} which ranked first on the latest CVPR'24 IMC Challenge in camera pose (rotation and translation) estimation.  
}%
\label{tab:IMCpose}
\end{table}

\section{Qualitative Examples}%
\label{sec:qual}

We further present qualitative examples of single-view reconstruction in \cref{fig:pointmap_supp}.

\section{Related Work}%
\label{sec:addrel}

In this section, we discuss additional related works.

\paragraph{Vision Transformers.} The Transformer architecture was initially proposed for language processing tasks~\cite{vaswani2017attention, Devlin2019BERTPO, brown2020language}. It was later introduced to the computer vision community by ViT~\cite{dosovitskiy21an-image}, sparking widespread adoption. Vision Transformers and their variants have since become dominant in the design of architectures for various computer vision tasks~\cite{arnab2021vivit, peebles2023scalable, cheng2022masked, xie2021segformer}, thanks to their simplicity, high capacity, flexibility, and ability to capture long-range dependencies.

DeiT~\cite{touvron2021training} demonstrated that Vision Transformers can be effectively trained on datasets like ImageNet using strong data augmentation strategies. DINO~\cite{caron21emerging} revealed intriguing properties of features learned by Vision Transformers in a self-supervised manner. CaiT~\cite{touvron2021going} introduced layer scaling to address the challenges of training deeper Vision Transformers, effectively mitigating gradient-related issues.
Further, techniques such as QKNorm~\cite{henry2020query, zhai2023stabilizing} have been proposed to stabilize the training process.
Additionally,~\cite{xie2022correlation} also explores the dynamics between frame-wise and global attention modules in object tracking, though using cross-attention.

\paragraph{Camera Pose Estimation.} 
Estimating camera poses from multi-view images is a crucial problem in 3D computer vision. 
Over the last decades, Structure from Motion (SfM) has emerged as the dominant approach~\cite{hartley04multiple}, whether incremental~\cite{snavely2006photo,agarwal2011building, frahm2010building,wu2013towards, schoenberger2016sfm} or global~\cite{arie2012global, crandall2012sfm, cui2015linear, moulon2013global, sweeney2015optimizing, rother2003linear, ozyesil2015robust, jiang2013global, cui2015global, cui2017hsfm,pan2024glomap}. 
Recently, a set of methods treat camera pose estimation as a regression problem~\cite{zhou2017unsupervised,ummenhofer2017demon,tang2018ba,wei2020deepsfm,wang2021deep, teed2018deepv2d, teed2021droid, zhang2022relpose,sinha2023sparsepose, wang2023posediffusion, lin2023relposepp, raydiffusion}, which show promising results under the sparse-view setting. 
Ace-Zero~\cite{brachmann2024acezero} further proposes to regress 3D scene coordinates and FlowMap~\cite{smith2024flowmap} focuses on depth maps, as intermediates for camera prediction. 
Instead, VGGSfM~\cite{wang24vggsfm:} simplifies the classical SfM pipeline to a differentiable framework, demonstrating exceptional performance, particularly with phototourism datasets. 
At the same time,  DUSt3R~\cite{wang24dust3r:, mast3r} introduces an approach to learn pixel-aligned point map, and hence camera poses can be recovered by simple alignment. 
This paradigm shift has garnered considerable interest as the point map, an over-parameterized representation, offers seamless integration with various downstream applications, such as 3D Gaussian splatting.


%
\newpage
{
\small
\bibliographystyle{ieeenat_fullname}
\bibliography{main,refs,vedaldi_general,vedaldi_specific}

\begin{thebibliography}{161}
\providecommand{\natexlab}[1]{#1}
\providecommand{\url}[1]{\texttt{#1}}
\expandafter\ifx\csname urlstyle\endcsname\relax
  \providecommand{\doi}[1]{doi: #1}\else
  \providecommand{\doi}{doi: \begingroup \urlstyle{rm}\Url}\fi

\bibitem[Achiam et~al.(2023)Achiam, Adler, Agarwal, Ahmad, Akkaya, Aleman, Almeida, Altenschmidt, Altman, Anadkat, et~al.]{achiam2023gpt}
Josh Achiam, Steven Adler, Sandhini Agarwal, Lama Ahmad, Ilge Akkaya, Florencia~Leoni Aleman, Diogo Almeida, Janko Altenschmidt, Sam Altman, Shyamal Anadkat, et~al.
\newblock Gpt-4 technical report.
\newblock \emph{arXiv preprint arXiv:2303.08774}, 2023.

\bibitem[Agarwal et~al.(2011)Agarwal, Furukawa, Snavely, Simon, Curless, Seitz, and Szeliski]{agarwal2011building}
Sameer Agarwal, Yasutaka Furukawa, Noah Snavely, Ian Simon, Brian Curless, Steven~M Seitz, and Richard Szeliski.
\newblock Building rome in a day.
\newblock \emph{Communications of the ACM}, 54\penalty0 (10):\penalty0 105--112, 2011.

\bibitem[Arie-Nachimson et~al.(2012)Arie-Nachimson, Kovalsky, Kemelmacher-Shlizerman, Singer, and Basri]{arie2012global}
Mica Arie-Nachimson, Shahar~Z Kovalsky, Ira Kemelmacher-Shlizerman, Amit Singer, and Ronen Basri.
\newblock Global motion estimation from point matches.
\newblock In \emph{2012 Second international conference on 3D imaging, modeling, processing, visualization \& transmission}, pages 81--88. IEEE, 2012.

\bibitem[Arnab et~al.(2021)Arnab, Dehghani, Heigold, Sun, Lu{\v{c}}i{\'c}, and Schmid]{arnab2021vivit}
Anurag Arnab, Mostafa Dehghani, Georg Heigold, Chen Sun, Mario Lu{\v{c}}i{\'c}, and Cordelia Schmid.
\newblock Vivit: A video vision transformer.
\newblock In \emph{Proceedings of the IEEE/CVF international conference on computer vision}, pages 6836--6846, 2021.

\bibitem[Brachmann et~al.(2024)Brachmann, Wynn, Chen, Cavallari, Monszpart, Turmukhambetov, and Prisacariu]{brachmann2024acezero}
Eric Brachmann, Jamie Wynn, Shuai Chen, Tommaso Cavallari, {\'{A}}ron Monszpart, Daniyar Turmukhambetov, and Victor~Adrian Prisacariu.
\newblock Scene coordinate reconstruction: Posing of image collections via incremental learning of a relocalizer.
\newblock In \emph{ECCV}, 2024.

\bibitem[Brown(2020)]{brown2020language}
Tom~B Brown.
\newblock Language models are few-shot learners.
\newblock \emph{arXiv preprint arXiv:2005.14165}, 2020.

\bibitem[Cabon et~al.(2020)Cabon, Murray, and Humenberger]{cabon2020virtual}
Yohann Cabon, Naila Murray, and Martin Humenberger.
\newblock Virtual kitti 2.
\newblock \emph{arXiv preprint arXiv:2001.10773}, 2020.

\bibitem[Cao et~al.(2025)Cao, Johnson, Vedaldi, and Novotny]{cao25lightplane:}
Ang Cao, Justin Johnson, Andrea Vedaldi, and David Novotny.
\newblock Lightplane: Highly-scalable components for neural {3D}fields.
\newblock In \emph{Proceedings of the International Conference on {3D} Vision (3DV)}, 2025.

\bibitem[Cao and Fu(2023)]{cao2023casmtr}
Chenjie Cao and Yanwei Fu.
\newblock Improving transformer-based image matching by cascaded capturing spatially informative keypoints.
\newblock In \emph{Proceedings of the IEEE/CVF International Conference on Computer Vision (ICCV)}, pages 12129--12139, 2023.

\bibitem[Caron et~al.(2021)Caron, Touvron, Misra, J{\'{e}}gou, Mairal, Bojanowski, and Joulin]{caron21emerging}
Mathilde Caron, Hugo Touvron, Ishan Misra, Herv{\'{e}} J{\'{e}}gou, Julien Mairal, Piotr Bojanowski, and Armand Joulin.
\newblock Emerging properties in self-supervised vision transformers.
\newblock In \emph{Proc. {ICCV}}, 2021.

\bibitem[Chen et~al.(2021)Chen, Luo, Zhang, Zhou, Bai, Hu, Tai, and Quan]{chen2021learning}
Hongkai Chen, Zixin Luo, Jiahui Zhang, Lei Zhou, Xuyang Bai, Zeyu Hu, Chiew-Lan Tai, and Long Quan.
\newblock Learning to match features with seeded graph matching network.
\newblock In \emph{Proceedings of the IEEE/CVF International Conference on Computer Vision}, pages 6301--6310, 2021.

\bibitem[Cheng et~al.(2022)Cheng, Misra, Schwing, Kirillov, and Girdhar]{cheng2022masked}
Bowen Cheng, Ishan Misra, Alexander~G Schwing, Alexander Kirillov, and Rohit Girdhar.
\newblock Masked-attention mask transformer for universal image segmentation.
\newblock In \emph{Proceedings of the IEEE/CVF conference on computer vision and pattern recognition}, pages 1290--1299, 2022.

\bibitem[Cho et~al.(2024)Cho, Huang, Nam, An, Kim, and Lee]{cho2024local}
Seokju Cho, Jiahui Huang, Jisu Nam, Honggyu An, Seungryong Kim, and Joon-Young Lee.
\newblock Local all-pair correspondence for point tracking.
\newblock \emph{Proc. {ECCV}}, 2024.

\bibitem[Crandall et~al.(2012)Crandall, Owens, Snavely, and Huttenlocher]{crandall2012sfm}
David~J Crandall, Andrew Owens, Noah Snavely, and Daniel~P Huttenlocher.
\newblock Sfm with mrfs: Discrete-continuous optimization for large-scale structure from motion.
\newblock \emph{IEEE transactions on pattern analysis and machine intelligence}, 35\penalty0 (12):\penalty0 2841--2853, 2012.

\bibitem[Cui et~al.(2017)Cui, Gao, Shen, and Hu]{cui2017hsfm}
Hainan Cui, Xiang Gao, Shuhan Shen, and Zhanyi Hu.
\newblock Hsfm: Hybrid structure-from-motion.
\newblock In \emph{Proceedings of the IEEE conference on computer vision and pattern recognition}, pages 1212--1221, 2017.

\bibitem[Cui and Tan(2015)]{cui2015global}
Zhaopeng Cui and Ping Tan.
\newblock Global structure-from-motion by similarity averaging.
\newblock In \emph{Proceedings of the IEEE International Conference on Computer Vision}, pages 864--872, 2015.

\bibitem[Cui et~al.(2015)Cui, Jiang, Tang, and Tan]{cui2015linear}
Zhaopeng Cui, Nianjuan Jiang, Chengzhou Tang, and Ping Tan.
\newblock Linear global translation estimation with feature tracks.
\newblock \emph{arXiv preprint arXiv:1503.01832}, 2015.

\bibitem[Dai et~al.(2017)Dai, Chang, Savva, Halber, Funkhouser, and Nie{\ss}ner]{dai2017scannet}
Angela Dai, Angel~X Chang, Manolis Savva, Maciej Halber, Thomas Funkhouser, and Matthias Nie{\ss}ner.
\newblock Scannet: Richly-annotated 3d reconstructions of indoor scenes.
\newblock In \emph{Proceedings of the IEEE conference on computer vision and pattern recognition}, pages 5828--5839, 2017.

\bibitem[Darcet et~al.(2023)Darcet, Oquab, Mairal, and Bojanowski]{registers}
Timoth{\'e}e Darcet, Maxime Oquab, Julien Mairal, and Piotr Bojanowski.
\newblock Vision transformers need registers.
\newblock \emph{arXiv preprint arXiv:2309.16588}, 2023.

\bibitem[Deitke et~al.(2023)Deitke, Schwenk, Salvador, Weihs, Michel, VanderBilt, Schmidt, Ehsani, Kembhavi, and Farhadi]{deitke2023objaverse}
Matt Deitke, Dustin Schwenk, Jordi Salvador, Luca Weihs, Oscar Michel, Eli VanderBilt, Ludwig Schmidt, Kiana Ehsani, Aniruddha Kembhavi, and Ali Farhadi.
\newblock Objaverse: A universe of annotated 3d objects.
\newblock In \emph{Proceedings of the IEEE/CVF Conference on Computer Vision and Pattern Recognition}, pages 13142--13153, 2023.

\bibitem[DeTone et~al.(2018)DeTone, Malisiewicz, and Rabinovich]{detone2018self}
Daniel DeTone, Tomasz Malisiewicz, and Andrew Rabinovich.
\newblock Superpoint: Self-supervised interest point detection and description.
\newblock In \emph{Proceedings of the IEEE conference on computer vision and pattern recognition workshops}, pages 224--236, 2018.

\bibitem[Devlin et~al.(2019)Devlin, Chang, Lee, and Toutanova]{Devlin2019BERTPO}
Jacob Devlin, Ming-Wei Chang, Kenton Lee, and Kristina Toutanova.
\newblock Bert: Pre-training of deep bidirectional transformers for language understanding.
\newblock In \emph{North American Chapter of the Association for Computational Linguistics}, 2019.

\bibitem[Doersch et~al.(2022)Doersch, Gupta, Markeeva, Recasens, Smaira, Aytar, Carreira, Zisserman, and Yang]{tapvid}
Carl Doersch, Ankush Gupta, Larisa Markeeva, Adri{\`a} Recasens, Lucas Smaira, Yusuf Aytar, Jo{\~a}o Carreira, Andrew Zisserman, and Yi Yang.
\newblock Tap-vid: A benchmark for tracking any point in a video.
\newblock \emph{arXiv}, 2022.

\bibitem[Doersch et~al.(2023{\natexlab{a}})Doersch, Yang, Vecerik, Gokay, Gupta, Aytar, Carreira, and Zisserman]{doersch2023tapir}
Carl Doersch, Yi Yang, Mel Vecerik, Dilara Gokay, Ankush Gupta, Yusuf Aytar, Joao Carreira, and Andrew Zisserman.
\newblock {TAPIR}: Tracking any point with per-frame initialization and temporal refinement.
\newblock \emph{arXiv}, 2306.08637, 2023{\natexlab{a}}.

\bibitem[Doersch et~al.(2023{\natexlab{b}})Doersch, Yang, Vecerik, Gokay, Gupta, Aytar, Carreira, and Zisserman]{doersch23tapir:}
Carl Doersch, Yi Yang, Mel Vecerik, Dilara Gokay, Ankush Gupta, Yusuf Aytar, Joao Carreira, and Andrew Zisserman.
\newblock {TAPIR:} tracking any point with per-frame initialization and temporal refinement.
\newblock In \emph{Proc. {CVPR}}, 2023{\natexlab{b}}.

\bibitem[Doersch et~al.(2024)Doersch, Yang, Gokay, Luc, Koppula, Gupta, Heyward, Goroshin, Carreira, and Zisserman]{doersch2024bootstap}
Carl Doersch, Yi Yang, Dilara Gokay, Pauline Luc, Skanda Koppula, Ankush Gupta, Joseph Heyward, Ross Goroshin, Jo{\~a}o Carreira, and Andrew Zisserman.
\newblock Bootstap: Bootstrapped training for tracking-any-point.
\newblock \emph{arXiv preprint arXiv:2402.00847}, 2024.

\bibitem[Dosovitskiy et~al.(2021)Dosovitskiy, Beyer, Kolesnikov, Weissenborn, Zhai, Unterthiner, Dehghani, Minderer, Heigold, Gelly, Uszkoreit, and Houlsby]{dosovitskiy21an-image}
Alexey Dosovitskiy, Lucas Beyer, Alexander Kolesnikov, Dirk Weissenborn, Xiaohua Zhai, Thomas Unterthiner, Mostafa Dehghani, Matthias Minderer, Georg Heigold, Sylvain Gelly, Jakob Uszkoreit, and Neil Houlsby.
\newblock An image is worth 16$\times$16 words: Transformers for image recognition at scale.
\newblock In \emph{Proc. {ICLR}}, 2021.

\bibitem[Downs et~al.(2022)Downs, Francis, Koenig, Kinman, Hickman, Reymann, McHugh, and Vanhoucke]{downs2022google}
Laura Downs, Anthony Francis, Nate Koenig, Brandon Kinman, Ryan Hickman, Krista Reymann, Thomas~B McHugh, and Vincent Vanhoucke.
\newblock Google scanned objects: A high-quality dataset of 3d scanned household items.
\newblock In \emph{2022 International Conference on Robotics and Automation (ICRA)}, pages 2553--2560. IEEE, 2022.

\bibitem[Dubey et~al.(2024)Dubey, Jauhri, Pandey, Kadian, Al{-}Dahle, Letman, Mathur, Schelten, Yang, Fan, Goyal, Hartshorn, Yang, Mitra, Sravankumar, Korenev, Hinsvark, Rao, Zhang, Rodriguez, Gregerson, Spataru, Rozi{\`{e}}re, Biron, Tang, Chern, Caucheteux, Nayak, Bi, Marra, McConnell, Keller, Touret, Wu, Wong, Ferrer, Nikolaidis, Allonsius, Song, Pintz, Livshits, Esiobu, Choudhary, Mahajan, Garcia{-}Olano, Perino, Hupkes, Lakomkin, AlBadawy, Lobanova, Dinan, Smith, Radenovic, Zhang, Synnaeve, Lee, Anderson, Nail, Mialon, Pang, Cucurell, Nguyen, Korevaar, Xu, Touvron, Zarov, Ibarra, Kloumann, Misra, Evtimov, Copet, Lee, Geffert, Vranes, Park, Mahadeokar, Shah, van~der Linde, Billock, Hong, Lee, Fu, Chi, Huang, Liu, Wang, Yu, Bitton, Spisak, Park, Rocca, Johnstun, Saxe, Jia, Alwala, Upasani, Plawiak, Li, Heafield, and Stone]{dubey24the-llama}
Abhimanyu Dubey, Abhinav Jauhri, Abhinav Pandey, Abhishek Kadian, Ahmad Al{-}Dahle, Aiesha Letman, Akhil Mathur, Alan Schelten, Amy Yang, Angela Fan, Anirudh Goyal, Anthony Hartshorn, Aobo Yang, Archi Mitra, Archie Sravankumar, Artem Korenev, Arthur Hinsvark, Arun Rao, Aston Zhang, Aur{\'{e}}lien Rodriguez, Austen Gregerson, Ava Spataru, Baptiste Rozi{\`{e}}re, Bethany Biron, Binh Tang, Bobbie Chern, Charlotte Caucheteux, Chaya Nayak, Chloe Bi, Chris Marra, Chris McConnell, Christian Keller, Christophe Touret, Chunyang Wu, Corinne Wong, Cristian~Canton Ferrer, Cyrus Nikolaidis, Damien Allonsius, Daniel Song, Danielle Pintz, Danny Livshits, David Esiobu, Dhruv Choudhary, Dhruv Mahajan, Diego Garcia{-}Olano, Diego Perino, Dieuwke Hupkes, Egor Lakomkin, Ehab AlBadawy, Elina Lobanova, Emily Dinan, Eric~Michael Smith, Filip Radenovic, Frank Zhang, Gabriel Synnaeve, Gabrielle Lee, Georgia~Lewis Anderson, Graeme Nail, Gr{\'{e}}goire Mialon, Guan Pang, Guillem Cucurell, Hailey Nguyen, Hannah Korevaar, Hu Xu, Hugo
  Touvron, Iliyan Zarov, Imanol~Arrieta Ibarra, Isabel~M. Kloumann, Ishan Misra, Ivan Evtimov, Jade Copet, Jaewon Lee, Jan Geffert, Jana Vranes, Jason Park, Jay Mahadeokar, Jeet Shah, Jelmer van~der Linde, Jennifer Billock, Jenny Hong, Jenya Lee, Jeremy Fu, Jianfeng Chi, Jianyu Huang, Jiawen Liu, Jie Wang, Jiecao Yu, Joanna Bitton, Joe Spisak, Jongsoo Park, Joseph Rocca, Joshua Johnstun, Joshua Saxe, Junteng Jia, Kalyan~Vasuden Alwala, Kartikeya Upasani, Kate Plawiak, Ke Li, Kenneth Heafield, and Kevin Stone.
\newblock The {Llama} 3 herd of models.
\newblock \emph{arXiv}, 2407.21783, 2024.

\bibitem[Duisterhof et~al.(2024)Duisterhof, Zust, Weinzaepfel, Leroy, Cabon, and Revaud]{duisterhof24mast3r-sfm:}
Bardienus Duisterhof, Lojze Zust, Philippe Weinzaepfel, Vincent Leroy, Yohann Cabon, and Jerome Revaud.
\newblock {MASt3R-SfM:} a fully-integrated solution for unconstrained structure-from-motion.
\newblock \emph{arXiv}, 2409.19152, 2024.

\bibitem[Dusmanu et~al.(2019)Dusmanu, Rocco, Pajdla, Pollefeys, Sivic, Torii, and Sattler]{dusmanu2019d2}
Mihai Dusmanu, Ignacio Rocco, Tomas Pajdla, Marc Pollefeys, Josef Sivic, Akihiko Torii, and Torsten Sattler.
\newblock D2-net: A trainable cnn for joint description and detection of local features.
\newblock In \emph{Proceedings of the ieee/cvf conference on computer vision and pattern recognition}, pages 8092--8101, 2019.

\bibitem[Edstedt et~al.(2023)Edstedt, Athanasiadis, Wadenbäck, and Felsberg]{edstedt2023dkm}
Johan Edstedt, Ioannis Athanasiadis, Mårten Wadenbäck, and Michael Felsberg.
\newblock {DKM}: Dense kernelized feature matching for geometry estimation.
\newblock In \emph{IEEE Conference on Computer Vision and Pattern Recognition}, 2023.

\bibitem[Edstedt et~al.(2024)Edstedt, Sun, B{\"o}kman, Wadenb{\"a}ck, and Felsberg]{edstedt2024roma}
Johan Edstedt, Qiyu Sun, Georg B{\"o}kman, M{\aa}rten Wadenb{\"a}ck, and Michael Felsberg.
\newblock Roma: Robust dense feature matching.
\newblock In \emph{Proceedings of the IEEE/CVF Conference on Computer Vision and Pattern Recognition}, pages 19790--19800, 2024.

\bibitem[Esser et~al.(2021)Esser, Rombach, and Ommer]{esser21taming}
Patrick Esser, Robin Rombach, and Bj{\"{o}}rn Ommer.
\newblock Taming transformers for high-resolution image synthesis.
\newblock In \emph{Proc. {CVPR}}, 2021.

\bibitem[Fischler and Bolles(1981)]{fischler1981random}
Martin~A Fischler and Robert~C Bolles.
\newblock Random sample consensus: a paradigm for model fitting with applications to image analysis and automated cartography.
\newblock \emph{Communications of the ACM}, 24\penalty0 (6):\penalty0 381--395, 1981.

\bibitem[Frahm et~al.(2010)Frahm, Fite-Georgel, Gallup, Johnson, Raguram, Wu, Jen, Dunn, Clipp, Lazebnik, et~al.]{frahm2010building}
Jan-Michael Frahm, Pierre Fite-Georgel, David Gallup, Tim Johnson, Rahul Raguram, Changchang Wu, Yi-Hung Jen, Enrique Dunn, Brian Clipp, Svetlana Lazebnik, et~al.
\newblock Building rome on a cloudless day.
\newblock In \emph{Computer Vision--ECCV 2010: 11th European Conference on Computer Vision, Heraklion, Crete, Greece, September 5-11, 2010, Proceedings, Part IV 11}, pages 368--381. Springer, 2010.

\bibitem[Fu et~al.(2022)Fu, Xu, Ong, and Tao]{fu2022geo}
Qiancheng Fu, Qingshan Xu, Yew~Soon Ong, and Wenbing Tao.
\newblock Geo-neus: Geometry-consistent neural implicit surfaces learning for multi-view reconstruction.
\newblock \emph{Advances in Neural Information Processing Systems}, 35:\penalty0 3403--3416, 2022.

\bibitem[Furukawa et~al.(2015)Furukawa, Hern{\'a}ndez, et~al.]{furukawa2015multi}
Yasutaka Furukawa, Carlos Hern{\'a}ndez, et~al.
\newblock Multi-view stereo: A tutorial.
\newblock \emph{Foundations and Trends{\textregistered} in Computer Graphics and Vision}, 9\penalty0 (1-2):\penalty0 1--148, 2015.

\bibitem[Galliani et~al.(2015{\natexlab{a}})Galliani, Lasinger, and Schindler]{galliani2015massively}
Silvano Galliani, Katrin Lasinger, and Konrad Schindler.
\newblock Massively parallel multiview stereopsis by surface normal diffusion.
\newblock In \emph{Proceedings of the IEEE international conference on computer vision}, pages 873--881, 2015{\natexlab{a}}.

\bibitem[Galliani et~al.(2015{\natexlab{b}})Galliani, Lasinger, and Schindler]{gipuma}
Silvano Galliani, Katrin Lasinger, and Konrad Schindler.
\newblock Massively parallel multiview stereopsis by surface normal diffusion.
\newblock In \emph{ICCV}, 2015{\natexlab{b}}.

\bibitem[Greff et~al.(2022)Greff, Belletti, Beyer, Doersch, Du, Duckworth, Fleet, Gnanapragasam, Golemo, Herrmann, Kipf, Kundu, Lagun, Laradji, Liu, Meyer, Miao, Nowrouzezahrai, Oztireli, Pot, Radwan, Rebain, Sabour, Sajjadi, Sela, Sitzmann, Stone, Sun, Vora, Wang, Wu, Yi, Zhong, and Tagliasacchi]{greff22kubric:}
Klaus Greff, Francois Belletti, Lucas Beyer, Carl Doersch, Yilun Du, Daniel Duckworth, David~J Fleet, Dan Gnanapragasam, Florian Golemo, Charles Herrmann, Thomas Kipf, Abhijit Kundu, Dmitry Lagun, Issam Laradji, Hsueh-Ti~(Derek) Liu, Henning Meyer, Yishu Miao, Derek Nowrouzezahrai, Cengiz Oztireli, Etienne Pot, Noha Radwan, Daniel Rebain, Sara Sabour, Mehdi S.~M. Sajjadi, Matan Sela, Vincent Sitzmann, Austin Stone, Deqing Sun, Suhani Vora, Ziyu Wang, Tianhao Wu, Kwang~Moo Yi, Fangcheng Zhong, and Andrea Tagliasacchi.
\newblock Kubric: a scalable dataset generator.
\newblock In \emph{Proc. {CVPR}}, 2022.

\bibitem[Gu et~al.(2020)Gu, Fan, Zhu, Dai, Tan, and Tan]{gu2020cascade}
Xiaodong Gu, Zhiwen Fan, Siyu Zhu, Zuozhuo Dai, Feitong Tan, and Ping Tan.
\newblock Cascade cost volume for high-resolution multi-view stereo and stereo matching.
\newblock In \emph{Proceedings of the IEEE/CVF conference on computer vision and pattern recognition}, pages 2495--2504, 2020.

\bibitem[Han et~al.(2024)Han, Wang, Vedaldi, Torr, and Kokkinos]{han2024flex3d}
Junlin Han, Jianyuan Wang, Andrea Vedaldi, Philip Torr, and Filippos Kokkinos.
\newblock Flex3d: Feed-forward 3d generation with flexible reconstruction model and input view curation.
\newblock \emph{arXiv preprint arXiv:2410.00890}, 2024.

\bibitem[Harley et~al.(2022)Harley, Fang, and Fragkiadaki]{pips}
Adam~W Harley, Zhaoyuan Fang, and Katerina Fragkiadaki.
\newblock Particle video revisited: Tracking through occlusions using point trajectories.
\newblock In \emph{Proc. {ECCV}}, 2022.

\bibitem[Hartley and Zisserman(2000)]{hartley_multiple_2000}
Richard Hartley and Andrew Zisserman.
\newblock \emph{Multiple {View} {Geometry} in {Computer} {Vision}}.
\newblock Cambridge University Press, 2000.

\bibitem[Hartley and Zisserman(2004)]{hartley04multiple}
Richard Hartley and Andrew Zisserman.
\newblock \emph{Multiple View Geometry in Computer Vision}.
\newblock Cambridge University Press, ISBN: 0521540518, 2004.

\bibitem[He et~al.(2023)He, Sun, Wang, Peng, Huang, Bao, and Zhou]{he2023dfsfm}
Xingyi He, Jiaming Sun, Yifan Wang, Sida Peng, Qixing Huang, Hujun Bao, and Xiaowei Zhou.
\newblock Detector-free structure from motion.
\newblock In \emph{arxiv}, 2023.

\bibitem[Henry et~al.(2020)Henry, Dachapally, Pawar, and Chen]{henry2020query}
Alex Henry, Prudhvi~Raj Dachapally, Shubham Pawar, and Yuxuan Chen.
\newblock Query-key normalization for transformers.
\newblock \emph{arXiv preprint arXiv:2010.04245}, 2020.

\bibitem[Hong et~al.(2024)Hong, Zhang, Gu, Bi, Zhou, Liu, Liu, Sunkavalli, Bui, and Tan]{hong24lrm:}
Yicong Hong, Kai Zhang, Jiuxiang Gu, Sai Bi, Yang Zhou, Difan Liu, Feng Liu, Kalyan Sunkavalli, Trung Bui, and Hao Tan.
\newblock {LRM}: Large reconstruction model for single image to {3D}.
\newblock In \emph{Proc. {ICLR}}, 2024.

\bibitem[Huang et~al.(2018)Huang, Matzen, Kopf, Ahuja, and Huang]{mvssynth}
Po-Han Huang, Kevin Matzen, Johannes Kopf, Narendra Ahuja, and Jia-Bin Huang.
\newblock Deepmvs: Learning multi-view stereopsis.
\newblock In \emph{IEEE Conference on Computer Vision and Pattern Recognition (CVPR)}, 2018.

\bibitem[Jensen et~al.(2014)Jensen, Dahl, Vogiatzis, Tola, and Aan{\ae}s]{dtudataset}
Rasmus Jensen, Anders Dahl, George Vogiatzis, Engil Tola, and Henrik Aan{\ae}s.
\newblock Large scale multi-view stereopsis evaluation.
\newblock In \emph{2014 IEEE Conference on Computer Vision and Pattern Recognition}, pages 406--413. IEEE, 2014.

\bibitem[Jiang et~al.(2013)Jiang, Cui, and Tan]{jiang2013global}
Nianjuan Jiang, Zhaopeng Cui, and Ping Tan.
\newblock A global linear method for camera pose registration.
\newblock In \emph{Proceedings of the IEEE international conference on computer vision}, pages 481--488, 2013.

\bibitem[Jin et~al.(2024)Jin, Jiang, Tan, Zhang, Bi, Zhang, Luan, Snavely, and Xu]{jin24lvsm:}
Haian Jin, Hanwen Jiang, Hao Tan, Kai Zhang, Sai Bi, Tianyuan Zhang, Fujun Luan, Noah Snavely, and Zexiang Xu.
\newblock {LVSM:} a large view synthesis model with minimal {3D} inductive bias.
\newblock \emph{arXiv}, 2410.17242, 2024.

\bibitem[Jin et~al.(2021)Jin, Mishkin, Mishchuk, Matas, Fua, Yi, and Trulls]{jin2021image}
Yuhe Jin, Dmytro Mishkin, Anastasiia Mishchuk, Jiri Matas, Pascal Fua, Kwang~Moo Yi, and Eduard Trulls.
\newblock Image matching across wide baselines: From paper to practice.
\newblock \emph{International Journal of Computer Vision}, 129\penalty0 (2):\penalty0 517--547, 2021.

\bibitem[Karaev et~al.(2024{\natexlab{a}})Karaev, Makarov, Wang, Neverova, Vedaldi, and Rupprecht]{karaev2024cotracker3}
Nikita Karaev, Iurii Makarov, Jianyuan Wang, Natalia Neverova, Andrea Vedaldi, and Christian Rupprecht.
\newblock Cotracker3: Simpler and better point tracking by pseudo-labelling real videos.
\newblock \emph{arXiv preprint arXiv:2410.11831}, 2024{\natexlab{a}}.

\bibitem[Karaev et~al.(2024{\natexlab{b}})Karaev, Rocco, Graham, Neverova, Vedaldi, and Rupprecht]{karaev2023cotracker}
Nikita Karaev, Ignacio Rocco, Benjamin Graham, Natalia Neverova, Andrea Vedaldi, and Christian Rupprecht.
\newblock Cotracker: It is better to track together.
\newblock \emph{Proc. {ECCV}}, 2024{\natexlab{b}}.

\bibitem[Karaev et~al.(2024{\natexlab{c}})Karaev, Rocco, Graham, Neverova, Vedaldi, and Rupprecht]{karaev24cotracker}
Nikita Karaev, Ignacio Rocco, Ben Graham, Natalia Neverova, Andrea Vedaldi, and Christian Rupprecht.
\newblock {CoTracker}: It is better to track together.
\newblock In \emph{Proceedings of the European Conference on Computer Vision ({ECCV})}, 2024{\natexlab{c}}.

\bibitem[Kendall and Cipolla(2016)]{kendall16modelling}
Alex Kendall and Roberto Cipolla.
\newblock Modelling uncertainty in deep learning for camera relocalization.
\newblock In \emph{Proc. {ICRA}}. IEEE, 2016.

\bibitem[Kendall and Gal(2017)]{kendall17what}
Alex Kendall and Yarin Gal.
\newblock What uncertainties do we need in {Bayesian} deep learning for computer vision?
\newblock \emph{Proc. {NeurIPS}}, 2017.

\bibitem[Le~Moing et~al.(2024)Le~Moing, Ponce, and Schmid]{lemoing2024dense}
Guillaume Le~Moing, Jean Ponce, and Cordelia Schmid.
\newblock Dense optical tracking: Connecting the dots.
\newblock In \emph{CVPR}, 2024.

\bibitem[Lepetit et~al.(2009)Lepetit, Moreno-Noguer, and Fua]{lepetit2009ep}
Vincent Lepetit, Francesc Moreno-Noguer, and Pascal Fua.
\newblock Ep n p: An accurate o (n) solution to the p n p problem.
\newblock \emph{International journal of computer vision}, 81:\penalty0 155--166, 2009.

\bibitem[Leroy et~al.(2024)Leroy, Cabon, and Revaud]{mast3r}
Vincent Leroy, Yohann Cabon, and J{\'e}r{\^o}me Revaud.
\newblock Grounding image matching in 3d with mast3r.
\newblock \emph{arXiv preprint arXiv:2406.09756}, 2024.

\bibitem[Li et~al.(2024)Li, Zhang, Liu, Zeng, Ren, Li, and Zhang]{li2024taptr}
Hongyang Li, Hao Zhang, Shilong Liu, Zhaoyang Zeng, Tianhe Ren, Feng Li, and Lei Zhang.
\newblock Taptr: Tracking any point with transformers as detection.
\newblock \emph{arXiv preprint arXiv:2403.13042}, 2024.

\bibitem[Li and Snavely(2018)]{li2018megadepth}
Zhengqi Li and Noah Snavely.
\newblock Megadepth: Learning single-view depth prediction from internet photos.
\newblock In \emph{Proceedings of the IEEE conference on computer vision and pattern recognition}, pages 2041--2050, 2018.

\bibitem[Lin et~al.(2023)Lin, Zhang, Ramanan, and Tulsiani]{lin2023relposepp}
Amy Lin, Jason~Y Zhang, Deva Ramanan, and Shubham Tulsiani.
\newblock Relpose++: Recovering 6d poses from sparse-view observations.
\newblock \emph{arXiv preprint arXiv:2305.04926}, 2023.

\bibitem[Lindenberger et~al.(2021)Lindenberger, Sarlin, Larsson, and Pollefeys]{lindenberger21pixel-perfect}
Philipp Lindenberger, Paul{-}Edouard Sarlin, Viktor Larsson, and Marc Pollefeys.
\newblock Pixel-perfect structure-from-motion with featuremetric refinement.
\newblock \emph{arXiv.cs}, abs/2108.08291, 2021.

\bibitem[Lindenberger et~al.(2023{\natexlab{a}})Lindenberger, Sarlin, and Pollefeys]{lindenberger2023lightglue}
Philipp Lindenberger, Paul-Edouard Sarlin, and Marc Pollefeys.
\newblock Lightglue: Local feature matching at light speed.
\newblock \emph{arXiv preprint arXiv:2306.13643}, 2023{\natexlab{a}}.

\bibitem[Lindenberger et~al.(2023{\natexlab{b}})Lindenberger, Sarlin, and Pollefeys]{lindenberger23lightglue:}
Philipp Lindenberger, Paul-Edouard Sarlin, and Marc Pollefeys.
\newblock {LightGlue:} local feature matching at light speed.
\newblock In \emph{Proc. {ICCV}}, 2023{\natexlab{b}}.

\bibitem[Ling et~al.(2024)Ling, Sheng, Tu, Zhao, Xin, Wan, Yu, Guo, Yu, Lu, et~al.]{ling2024dl3dv}
Lu Ling, Yichen Sheng, Zhi Tu, Wentian Zhao, Cheng Xin, Kun Wan, Lantao Yu, Qianyu Guo, Zixun Yu, Yawen Lu, et~al.
\newblock Dl3dv-10k: A large-scale scene dataset for deep learning-based 3d vision.
\newblock In \emph{Proceedings of the IEEE/CVF Conference on Computer Vision and Pattern Recognition}, pages 22160--22169, 2024.

\bibitem[Liu et~al.(2025)Liu, Gao, Zhang, Pautrat, Sch{\"o}nberger, Larsson, and Pollefeys]{liu2025robust}
Shaohui Liu, Yidan Gao, Tianyi Zhang, R{\'e}mi Pautrat, Johannes~L Sch{\"o}nberger, Viktor Larsson, and Marc Pollefeys.
\newblock Robust incremental structure-from-motion with hybrid features.
\newblock In \emph{European Conference on Computer Vision}, pages 249--269. Springer, 2025.

\bibitem[Lopez-Antequera et~al.(2020)Lopez-Antequera, Gargallo, Hofinger, Rota~BulÃ², Kuang, and Kontschieder]{MPSD_2020_ECCV}
Manuel Lopez-Antequera, Pau Gargallo, Markus Hofinger, Samuel Rota~BulÃ², Yubin Kuang, and Peter Kontschieder.
\newblock Mapillary planet-scale depth dataset.
\newblock In \emph{Proceedings of the European Conference on Computer Vision (ECCV)}, 2020.

\bibitem[Ma et~al.(2022)Ma, Teed, and Deng]{ma2022multiview}
Zeyu Ma, Zachary Teed, and Jia Deng.
\newblock Multiview stereo with cascaded epipolar raft.
\newblock In \emph{European Conference on Computer Vision}, pages 734--750. Springer, 2022.

\bibitem[Moulon et~al.(2013)Moulon, Monasse, and Marlet]{moulon2013global}
Pierre Moulon, Pascal Monasse, and Renaud Marlet.
\newblock Global fusion of relative motions for robust, accurate and scalable structure from motion.
\newblock In \emph{Proceedings of the IEEE international conference on computer vision}, pages 3248--3255, 2013.

\bibitem[Niemeyer et~al.(2020)Niemeyer, Mescheder, Oechsle, and Geiger]{niemeyer2020differentiable}
Michael Niemeyer, Lars Mescheder, Michael Oechsle, and Andreas Geiger.
\newblock Differentiable volumetric rendering: Learning implicit 3d representations without 3d supervision.
\newblock In \emph{Proceedings of the IEEE/CVF conference on computer vision and pattern recognition}, pages 3504--3515, 2020.

\bibitem[Novotn{\'{y}} et~al.(2017)Novotn{\'{y}}, Larlus, and Vedaldi]{novotny17learning}
David Novotn{\'{y}}, Diane Larlus, and Andrea Vedaldi.
\newblock Learning {3D} object categories by looking around them.
\newblock In \emph{Proceedings of the International Conference on Computer Vision ({ICCV})}, 2017.

\bibitem[Novotn{\'{y}} et~al.(2018)Novotn{\'{y}}, Larlus, and Vedaldi]{novotny18capturing}
David Novotn{\'{y}}, Diane Larlus, and Andrea Vedaldi.
\newblock Capturing the geometry of object categories from video supervision.
\newblock \emph{{IEEE} Transactions on Pattern Analysis and Machine Intelligence}, 2018.

\bibitem[Oliensis(2000)]{oliensis2000critique}
John Oliensis.
\newblock A critique of structure-from-motion algorithms.
\newblock \emph{Computer Vision and Image Understanding}, 80\penalty0 (2):\penalty0 172--214, 2000.

\bibitem[Oquab et~al.(2024)Oquab, Darcet, Moutakanni, Vo, Szafraniec, Khalidov, Fernandez, HAZIZA, Massa, El-Nouby, Assran, Ballas, Galuba, Howes, Huang, Li, Misra, Rabbat, Sharma, Synnaeve, Xu, Jegou, Mairal, Labatut, Joulin, and Bojanowski]{oquab24dinov2:}
Maxime Oquab, Timoth{\'e}e Darcet, Th{\'e}o Moutakanni, Huy~V. Vo, Marc Szafraniec, Vasil Khalidov, Pierre Fernandez, Daniel HAZIZA, Francisco Massa, Alaaeldin El-Nouby, Mido Assran, Nicolas Ballas, Wojciech Galuba, Russell Howes, Po-Yao Huang, Shang-Wen Li, Ishan Misra, Michael Rabbat, Vasu Sharma, Gabriel Synnaeve, Hu Xu, Herve Jegou, Julien Mairal, Patrick Labatut, Armand Joulin, and Piotr Bojanowski.
\newblock {DINO}v2: Learning robust visual features without supervision.
\newblock \emph{Transactions on Machine Learning Research}, 2024.

\bibitem[Ozyesil and Singer(2015)]{ozyesil2015robust}
Onur Ozyesil and Amit Singer.
\newblock Robust camera location estimation by convex programming.
\newblock In \emph{Proceedings of the IEEE Conference on Computer Vision and Pattern Recognition}, pages 2674--2683, 2015.

\bibitem[{\"O}zye{\c{s}}il et~al.(2017){\"O}zye{\c{s}}il, Voroninski, Basri, and Singer]{ozyecsil2017survey}
Onur {\"O}zye{\c{s}}il, Vladislav Voroninski, Ronen Basri, and Amit Singer.
\newblock A survey of structure from motion*.
\newblock \emph{Acta Numerica}, 26:\penalty0 305--364, 2017.

\bibitem[Pan et~al.(2024)Pan, Barath, Pollefeys, and Sch\"{o}nberger]{pan2024glomap}
Linfei Pan, Daniel Barath, Marc Pollefeys, and Johannes~Lutz Sch\"{o}nberger.
\newblock {Global Structure-from-Motion Revisited}.
\newblock In \emph{European Conference on Computer Vision (ECCV)}, 2024.

\bibitem[Pan et~al.(2023)Pan, Charron, Yang, Peters, Whelan, Kong, Parkhi, Newcombe, and Ren]{Pan_2023_ICCV}
Xiaqing Pan, Nicholas Charron, Yongqian Yang, Scott Peters, Thomas Whelan, Chen Kong, Omkar Parkhi, Richard Newcombe, and Yuheng~(Carl) Ren.
\newblock Aria digital twin: A new benchmark dataset for egocentric 3d machine perception.
\newblock In \emph{Proceedings of the IEEE/CVF International Conference on Computer Vision (ICCV)}, pages 20133--20143, 2023.

\bibitem[Peebles and Xie(2023)]{peebles2023scalable}
William Peebles and Saining Xie.
\newblock Scalable diffusion models with transformers.
\newblock In \emph{Proceedings of the IEEE/CVF International Conference on Computer Vision}, pages 4195--4205, 2023.

\bibitem[Peng et~al.(2022)Peng, Wang, Wang, Lai, and Wang]{peng2022rethinking}
Rui Peng, Rongjie Wang, Zhenyu Wang, Yawen Lai, and Ronggang Wang.
\newblock Rethinking depth estimation for multi-view stereo: A unified representation.
\newblock In \emph{Proceedings of the IEEE/CVF conference on computer vision and pattern recognition}, pages 8645--8654, 2022.

\bibitem[Pineda et~al.(2022)Pineda, Fan, Monge, Venkataraman, Sodhi, Chen, Ortiz, DeTone, Wang, Anderson, et~al.]{pineda2022theseus}
Luis Pineda, Taosha Fan, Maurizio Monge, Shobha Venkataraman, Paloma Sodhi, Ricky~TQ Chen, Joseph Ortiz, Daniel DeTone, Austin Wang, Stuart Anderson, et~al.
\newblock Theseus: A library for differentiable nonlinear optimization.
\newblock \emph{Advances in Neural Information Processing Systems}, 35:\penalty0 3801--3818, 2022.

\bibitem[Radford et~al.(2021)Radford, Kim, Hallacy, Ramesh, Goh, Agarwal, Sastry, Askell, Mishkin, Clark, Krueger, and Sutskever]{radford21learning}
Alec Radford, Jong~Wook Kim, Chris Hallacy, Aditya Ramesh, Gabriel Goh, Sandhini Agarwal, Girish Sastry, Amanda Askell, Pamela Mishkin, Jack Clark, Gretchen Krueger, and Ilya Sutskever.
\newblock Learning transferable visual models from natural language supervision.
\newblock In \emph{Proc. {ICML}}, pages 8748--8763, 2021.

\bibitem[Ranftl et~al.(2021)Ranftl, Bochkovskiy, and Koltun]{ranftl21dpt}
Ren{\'e} Ranftl, Alexey Bochkovskiy, and Vladlen Koltun.
\newblock Vision transformers for dense prediction.
\newblock In \emph{Proceedings of the IEEE/CVF international conference on computer vision}, pages 12179--12188, 2021.

\bibitem[Reizenstein et~al.(2021)Reizenstein, Shapovalov, Henzler, Sbordone, Labatut, and Novotny]{reizenstein21common}
Jeremy Reizenstein, Roman Shapovalov, Philipp Henzler, Luca Sbordone, Patrick Labatut, and David Novotny.
\newblock {Common Objects in 3D}: Large-scale learning and evaluation of real-life {3D} category reconstruction.
\newblock In \emph{Proc. {ICCV}}, 2021.

\bibitem[Roberts et~al.(2021)Roberts, Ramapuram, Ranjan, Kumar, Bautista, Paczan, Webb, and Susskind]{hypersim}
Mike Roberts, Jason Ramapuram, Anurag Ranjan, Atulit Kumar, Miguel~Angel Bautista, Nathan Paczan, Russ Webb, and Joshua~M. Susskind.
\newblock {Hypersim}: {A} photorealistic synthetic dataset for holistic indoor scene understanding.
\newblock In \emph{International Conference on Computer Vision (ICCV) 2021}, 2021.

\bibitem[Rother(2003)]{rother2003linear}
Rother.
\newblock Linear multiview reconstruction of points, lines, planes and cameras using a reference plane.
\newblock In \emph{Proceedings Ninth IEEE International Conference on Computer Vision}, pages 1210--1217. IEEE, 2003.

\bibitem[Sand and Teller(2008)]{sand2008particle}
Peter Sand and Seth Teller.
\newblock Particle video: Long-range motion estimation using point trajectories.
\newblock \emph{{IJCV}}, 80, 2008.

\bibitem[Sarlin et~al.(2020{\natexlab{a}})Sarlin, DeTone, Malisiewicz, and Rabinovich]{sarlin2020superglue}
Paul-Edouard Sarlin, Daniel DeTone, Tomasz Malisiewicz, and Andrew Rabinovich.
\newblock Superglue: Learning feature matching with graph neural networks.
\newblock In \emph{Proceedings of the IEEE/CVF conference on computer vision and pattern recognition}, pages 4938--4947, 2020{\natexlab{a}}.

\bibitem[Sarlin et~al.(2020{\natexlab{b}})Sarlin, DeTone, Malisiewicz, and Rabinovich]{sarlin20superglue:}
Paul-Edouard Sarlin, Daniel DeTone, Tomasz Malisiewicz, and Andrew Rabinovich.
\newblock {SuperGlue:} learning feature matching with graph neural networks.
\newblock In \emph{Proc. {CVPR}}, 2020{\natexlab{b}}.

\bibitem[Sch\"{o}nberger and Frahm(2016{\natexlab{a}})]{schoenberger2016sfm}
Johannes~Lutz Sch\"{o}nberger and Jan-Michael Frahm.
\newblock Structure-from-motion revisited.
\newblock In \emph{Conference on Computer Vision and Pattern Recognition (CVPR)}, 2016{\natexlab{a}}.

\bibitem[Sch\"{o}nberger and Frahm(2016{\natexlab{b}})]{schonberger16structure-from-motion}
Johannes~Lutz Sch\"{o}nberger and Jan-Michael Frahm.
\newblock Structure-from-motion revisited.
\newblock In \emph{Proc. {CVPR}}, 2016{\natexlab{b}}.

\bibitem[Sch{\"o}nberger et~al.(2016)Sch{\"o}nberger, Zheng, Frahm, and Pollefeys]{schonberger2016pixelwise}
Johannes~L Sch{\"o}nberger, Enliang Zheng, Jan-Michael Frahm, and Marc Pollefeys.
\newblock Pixelwise view selection for unstructured multi-view stereo.
\newblock In \emph{Computer Vision--ECCV 2016: 14th European Conference, Amsterdam, The Netherlands, October 11-14, 2016, Proceedings, Part III 14}, pages 501--518. Springer, 2016.

\bibitem[Schops et~al.(2017)Schops, Schonberger, Galliani, Sattler, Schindler, Pollefeys, and Geiger]{eth3d}
Thomas Schops, Johannes~L Schonberger, Silvano Galliani, Torsten Sattler, Konrad Schindler, Marc Pollefeys, and Andreas Geiger.
\newblock A multi-view stereo benchmark with high-resolution images and multi-camera videos.
\newblock In \emph{Proceedings of the IEEE conference on computer vision and pattern recognition}, pages 3260--3269, 2017.

\bibitem[Shah et~al.(2024)Shah, Bikshandi, Zhang, Thakkar, Ramani, and Dao]{shah2024flashattention}
Jay Shah, Ganesh Bikshandi, Ying Zhang, Vijay Thakkar, Pradeep Ramani, and Tri Dao.
\newblock Flashattention-3: Fast and accurate attention with asynchrony and low-precision.
\newblock \emph{Advances in Neural Information Processing Systems}, 37:\penalty0 68658--68685, 2024.

\bibitem[Shi et~al.(2022)Shi, Cai, Shavit, Mu, Feng, and Zhang]{shi2022clustergnn}
Yan Shi, Jun-Xiong Cai, Yoli Shavit, Tai-Jiang Mu, Wensen Feng, and Kai Zhang.
\newblock Clustergnn: Cluster-based coarse-to-fine graph neural network for efficient feature matching.
\newblock In \emph{Proceedings of the IEEE/CVF Conference on Computer Vision and Pattern Recognition}, pages 12517--12526, 2022.

\bibitem[Sinha et~al.(2023)Sinha, Zhang, Tagliasacchi, Gilitschenski, and Lindell]{sinha2023sparsepose}
Samarth Sinha, Jason~Y Zhang, Andrea Tagliasacchi, Igor Gilitschenski, and David~B Lindell.
\newblock Sparsepose: Sparse-view camera pose regression and refinement.
\newblock In \emph{Proceedings of the IEEE/CVF Conference on Computer Vision and Pattern Recognition}, pages 21349--21359, 2023.

\bibitem[Smith et~al.(2024{\natexlab{a}})Smith, Charatan, Tewari, and Sitzmann]{smith2024flowmap}
Cameron Smith, David Charatan, Ayush Tewari, and Vincent Sitzmann.
\newblock Flowmap: High-quality camera poses, intrinsics, and depth via gradient descent.
\newblock \emph{arXiv preprint arXiv:2404.15259}, 2024{\natexlab{a}}.

\bibitem[Smith et~al.(2024{\natexlab{b}})Smith, Charatan, Tewari, and Sitzmann]{smith24flowmap:}
Cameron Smith, David Charatan, Ayush Tewari, and Vincent Sitzmann.
\newblock {FlowMap:} high-quality camera poses, intrinsics, and depth via gradient descent.
\newblock \emph{arXiv}, 2404.15259, 2024{\natexlab{b}}.

\bibitem[Snavely et~al.(2006)Snavely, Seitz, and Szeliski]{snavely2006photo}
Noah Snavely, Steven~M Seitz, and Richard Szeliski.
\newblock Photo tourism: exploring photo collections in 3d.
\newblock In \emph{ACM siggraph 2006 papers}, pages 835--846. 2006.

\bibitem[Straub et~al.(2019)Straub, Whelan, Ma, Chen, Wijmans, Green, Engel, Mur-Artal, Ren, Verma, et~al.]{straub2019replica}
Julian Straub, Thomas Whelan, Lingni Ma, Yufan Chen, Erik Wijmans, Simon Green, Jakob~J Engel, Raul Mur-Artal, Carl Ren, Shobhit Verma, et~al.
\newblock The replica dataset: A digital replica of indoor spaces.
\newblock \emph{arXiv preprint arXiv:1906.05797}, 2019.

\bibitem[Sun et~al.(2021)Sun, Shen, Wang, Bao, and Zhou]{sun2021loftr}
Jiaming Sun, Zehong Shen, Yuang Wang, Hujun Bao, and Xiaowei Zhou.
\newblock Loftr: Detector-free local feature matching with transformers.
\newblock In \emph{Proceedings of the IEEE/CVF conference on computer vision and pattern recognition}, pages 8922--8931, 2021.

\bibitem[Sweeney et~al.(2015)Sweeney, Sattler, Hollerer, Turk, and Pollefeys]{sweeney2015optimizing}
Chris Sweeney, Torsten Sattler, Tobias Hollerer, Matthew Turk, and Marc Pollefeys.
\newblock Optimizing the viewing graph for structure-from-motion.
\newblock In \emph{Proceedings of the IEEE international conference on computer vision}, pages 801--809, 2015.

\bibitem[Szot et~al.(2021)Szot, Clegg, Undersander, Wijmans, Zhao, Turner, Maestre, Mukadam, Chaplot, Maksymets, Gokaslan, Vondrus, Dharur, Meier, Galuba, Chang, Kira, Koltun, Malik, Savva, and Batra]{szot2021habitat}
Andrew Szot, Alex Clegg, Eric Undersander, Erik Wijmans, Yili Zhao, John Turner, Noah Maestre, Mustafa Mukadam, Devendra Chaplot, Oleksandr Maksymets, Aaron Gokaslan, Vladimir Vondrus, Sameer Dharur, Franziska Meier, Wojciech Galuba, Angel Chang, Zsolt Kira, Vladlen Koltun, Jitendra Malik, Manolis Savva, and Dhruv Batra.
\newblock Habitat 2.0: Training home assistants to rearrange their habitat.
\newblock In \emph{Advances in Neural Information Processing Systems (NeurIPS)}, 2021.

\bibitem[Szymanowicz et~al.(2024)Szymanowicz, Rupprecht, and Vedaldi]{szymanowicz2024splatter}
Stanislaw Szymanowicz, Chrisitian Rupprecht, and Andrea Vedaldi.
\newblock Splatter image: Ultra-fast single-view 3d reconstruction.
\newblock In \emph{Proceedings of the IEEE/CVF conference on computer vision and pattern recognition}, pages 10208--10217, 2024.

\bibitem[Tang and Tan(2018)]{tang2018ba}
Chengzhou Tang and Ping Tan.
\newblock Ba-net: Dense bundle adjustment network.
\newblock \emph{arXiv preprint arXiv:1806.04807}, 2018.

\bibitem[Tang et~al.(2024{\natexlab{a}})Tang, Chen, Chen, Wang, Zeng, and Liu]{tang2024lgm}
Jiaxiang Tang, Zhaoxi Chen, Xiaokang Chen, Tengfei Wang, Gang Zeng, and Ziwei Liu.
\newblock Lgm: Large multi-view gaussian model for high-resolution 3d content creation.
\newblock In \emph{European Conference on Computer Vision}, pages 1--18. Springer, 2024{\natexlab{a}}.

\bibitem[Tang et~al.(2024{\natexlab{b}})Tang, Fan, Wang, Xu, Ranjan, Schwing, and Yan]{tang2024mv}
Zhenggang Tang, Yuchen Fan, Dilin Wang, Hongyu Xu, Rakesh Ranjan, Alexander Schwing, and Zhicheng Yan.
\newblock Mv-dust3r+: Single-stage scene reconstruction from sparse views in 2 seconds.
\newblock \emph{arXiv preprint arXiv:2412.06974}, 2024{\natexlab{b}}.

\bibitem[Teed and Deng(2018)]{teed2018deepv2d}
Zachary Teed and Jia Deng.
\newblock Deepv2d: Video to depth with differentiable structure from motion.
\newblock \emph{arXiv preprint arXiv:1812.04605}, 2018.

\bibitem[Teed and Deng(2021)]{teed2021droid}
Zachary Teed and Jia Deng.
\newblock Droid-slam: Deep visual slam for monocular, stereo, and rgb-d cameras.
\newblock \emph{Advances in neural information processing systems}, 34:\penalty0 16558--16569, 2021.

\bibitem[Touvron et~al.(2021{\natexlab{a}})Touvron, Cord, Douze, Massa, Sablayrolles, and J{\'e}gou]{touvron2021training}
Hugo Touvron, Matthieu Cord, Matthijs Douze, Francisco Massa, Alexandre Sablayrolles, and Herv{\'e} J{\'e}gou.
\newblock Training data-efficient image transformers \& distillation through attention.
\newblock In \emph{International conference on machine learning}, pages 10347--10357. PMLR, 2021{\natexlab{a}}.

\bibitem[Touvron et~al.(2021{\natexlab{b}})Touvron, Cord, Sablayrolles, Synnaeve, and J{\'e}gou]{touvron2021going}
Hugo Touvron, Matthieu Cord, Alexandre Sablayrolles, Gabriel Synnaeve, and Herv{\'e} J{\'e}gou.
\newblock Going deeper with image transformers.
\newblock In \emph{Proceedings of the IEEE/CVF international conference on computer vision}, pages 32--42, 2021{\natexlab{b}}.

\bibitem[Tyszkiewicz et~al.(2020)Tyszkiewicz, Fua, and Trulls]{tyszkiewicz2020disk}
Micha{\l} Tyszkiewicz, Pascal Fua, and Eduard Trulls.
\newblock Disk: Learning local features with policy gradient.
\newblock \emph{Advances in Neural Information Processing Systems}, 33:\penalty0 14254--14265, 2020.

\bibitem[Umeyama(1991)]{umeyama91least-squares}
Shinji Umeyama.
\newblock Least-squares estimation of transformation parameters between two point patterns.
\newblock \emph{{IEEE} Trans. Pattern Anal. Mach. Intell.}, 13\penalty0 (4), 1991.

\bibitem[Ummenhofer et~al.(2017)Ummenhofer, Zhou, Uhrig, Mayer, Ilg, Dosovitskiy, and Brox]{ummenhofer2017demon}
Benjamin Ummenhofer, Huizhong Zhou, Jonas Uhrig, Nikolaus Mayer, Eddy Ilg, Alexey Dosovitskiy, and Thomas Brox.
\newblock Demon: Depth and motion network for learning monocular stereo.
\newblock In \emph{Proceedings of the IEEE conference on computer vision and pattern recognition}, pages 5038--5047, 2017.

\bibitem[Vaswani et~al.(2017{\natexlab{a}})Vaswani, Shazeer, Parmar, Uszkoreit, Jones, Gomez, Kaiser, and Polosukhin]{vaswani17attention}
Ashish Vaswani, Noam Shazeer, Niki Parmar, Jakob Uszkoreit, Llion Jones, Aidan~N. Gomez, Lukasz Kaiser, and Illia Polosukhin.
\newblock Attention is all you need.
\newblock In \emph{Proc. {NeurIPS}}, 2017{\natexlab{a}}.

\bibitem[Vaswani et~al.(2017{\natexlab{b}})Vaswani, Shazeer, Parmar, Uszkoreit, Jones, Gomez, Kaiser, and Polosukhin]{vaswani2017attention}
Ashish Vaswani, Noam Shazeer, Niki Parmar, Jakob Uszkoreit, Llion Jones, Aidan~N Gomez, {\L}ukasz Kaiser, and Illia Polosukhin.
\newblock Attention is all you need.
\newblock \emph{Advances in neural information processing systems}, 30, 2017{\natexlab{b}}.

\bibitem[Wang et~al.(2021{\natexlab{a}})Wang, Galliani, Vogel, Speciale, and Pollefeys]{pathcmatchnet}
Fangjinhua Wang, Silvano Galliani, Christoph Vogel, Pablo Speciale, and Marc Pollefeys.
\newblock Patchmatchnet: Learned multi-view patchmatch stereo.
\newblock In \emph{CVPR}, pages 14194--14203, 2021{\natexlab{a}}.

\bibitem[Wang et~al.(2021{\natexlab{b}})Wang, Zhong, Dai, Birchfield, Zhang, Smolyanskiy, and Li]{wang2021deep}
Jianyuan Wang, Yiran Zhong, Yuchao Dai, Stan Birchfield, Kaihao Zhang, Nikolai Smolyanskiy, and Hongdong Li.
\newblock Deep two-view structure-from-motion revisited.
\newblock In \emph{Proceedings of the IEEE/CVF conference on Computer Vision and Pattern Recognition}, pages 8953--8962, 2021{\natexlab{b}}.

\bibitem[Wang et~al.(2023{\natexlab{a}})Wang, Rupprecht, and Novotny]{wang2023posediffusion}
Jianyuan Wang, Christian Rupprecht, and David Novotny.
\newblock Posediffusion: Solving pose estimation via diffusion-aided bundle adjustment.
\newblock In \emph{Proceedings of the IEEE/CVF International Conference on Computer Vision}, pages 9773--9783, 2023{\natexlab{a}}.

\bibitem[Wang et~al.(2023{\natexlab{b}})Wang, Rupprecht, and Novotny]{wang23posediffusion:}
Jianyuan Wang, Christian Rupprecht, and David Novotny.
\newblock {PoseDiffusion:} solving pose estimation via diffusion-aided bundle adjustment.
\newblock In \emph{Proc. {ICCV}}, 2023{\natexlab{b}}.

\bibitem[Wang et~al.(2024{\natexlab{a}})Wang, Karaev, Rupprecht, and Novotny]{wang24vggsfm:}
Jianyuan Wang, Nikita Karaev, Christian Rupprecht, and David Novotny.
\newblock {VGGSfM:} visual geometry grounded deep structure from motion.
\newblock In \emph{Proc. {CVPR}}, 2024{\natexlab{a}}.

\bibitem[Wang et~al.(2023{\natexlab{c}})Wang, Tan, Bi, Xu, Luan, Sunkavalli, Wang, Xu, and Zhang]{wang23pf-lrm:}
Peng Wang, Hao Tan, Sai Bi, Yinghao Xu, Fujun Luan, Kalyan Sunkavalli, Wenping Wang, Zexiang Xu, and Kai Zhang.
\newblock {PF-LRM:} pose-free large reconstruction model for joint pose and shape prediction.
\newblock \emph{arXiv.cs}, abs/2311.12024, 2023{\natexlab{c}}.

\bibitem[Wang et~al.(2025)Wang, Zhang, Holynski, Efros, and Kanazawa]{cut3r}
Qianqian Wang, Yifei Zhang, Aleksander Holynski, Alexei~A. Efros, and Angjoo Kanazawa.
\newblock Continuous 3d perception model with persistent state, 2025.

\bibitem[Wang et~al.(2024{\natexlab{b}})Wang, Xu, Dai, Xiang, Deng, Tong, and Yang]{wang24moge:}
Ruicheng Wang, Sicheng Xu, Cassie Dai, Jianfeng Xiang, Yu Deng, Xin Tong, and Jiaolong Yang.
\newblock {MoGe:} unlocking accurate monocular geometry estimation for open-domain images with optimal training supervision.
\newblock \emph{arXiv}, 2410.19115, 2024{\natexlab{b}}.

\bibitem[Wang et~al.(2024{\natexlab{c}})Wang, Leroy, Cabon, Chidlovskii, and Revaud]{wang24dust3r:}
Shuzhe Wang, Vincent Leroy, Yohann Cabon, Boris Chidlovskii, and Jerome Revaud.
\newblock {DUSt3R}: Geometric {3D} vision made easy.
\newblock In \emph{Proc. {CVPR}}, 2024{\natexlab{c}}.

\bibitem[Wang et~al.(2023{\natexlab{d}})Wang, Zeng, Guan, Yang, Chen, Liu, Xu, and Luo]{Wang_2023_CVPR}
Yuesong Wang, Zhaojie Zeng, Tao Guan, Wei Yang, Zhuo Chen, Wenkai Liu, Luoyuan Xu, and Yawei Luo.
\newblock Adaptive patch deformation for textureless-resilient multi-view stereo.
\newblock In \emph{Proceedings of the IEEE/CVF Conference on Computer Vision and Pattern Recognition (CVPR)}, pages 1621--1630, 2023{\natexlab{d}}.

\bibitem[Wei et~al.(2020)Wei, Zhang, Li, Fu, and Xue]{wei2020deepsfm}
Xingkui Wei, Yinda Zhang, Zhuwen Li, Yanwei Fu, and Xiangyang Xue.
\newblock Deepsfm: Structure from motion via deep bundle adjustment.
\newblock In \emph{Computer Vision--ECCV 2020: 16th European Conference, Glasgow, UK, August 23--28, 2020, Proceedings, Part I 16}, pages 230--247. Springer, 2020.

\bibitem[Wei et~al.(2024)Wei, Zhang, Bi, Tan, Luan, Deschaintre, Sunkavalli, Su, and Xu]{wei24meshlrm:}
Xinyue Wei, Kai Zhang, Sai Bi, Hao Tan, Fujun Luan, Valentin Deschaintre, Kalyan Sunkavalli, Hao Su, and Zexiang Xu.
\newblock {MeshLRM:} large reconstruction model for high-quality mesh.
\newblock \emph{arXiv}, 2404.12385, 2024.

\bibitem[Wei et~al.(2021)Wei, Liu, Rao, Zhao, Lu, and Zhou]{Wei_2021_ICCV}
Yi Wei, Shaohui Liu, Yongming Rao, Wang Zhao, Jiwen Lu, and Jie Zhou.
\newblock Nerfingmvs: Guided optimization of neural radiance fields for indoor multi-view stereo.
\newblock In \emph{Proceedings of the IEEE/CVF International Conference on Computer Vision (ICCV)}, pages 5610--5619, 2021.

\bibitem[Wu(2013)]{wu2013towards}
Changchang Wu.
\newblock Towards linear-time incremental structure from motion.
\newblock In \emph{2013 International Conference on 3D Vision-3DV 2013}, pages 127--134. IEEE, 2013.

\bibitem[Xia et~al.(2024)Xia, Fu, Liu, and Wang]{xia2024rgbd}
Hongchi Xia, Yang Fu, Sifei Liu, and Xiaolong Wang.
\newblock Rgbd objects in the wild: Scaling real-world 3d object learning from rgb-d videos, 2024.

\bibitem[Xiao et~al.(2024)Xiao, Wang, Zhang, Xue, Peng, Shen, and Zhou]{xiao2024spatialtracker}
Yuxi Xiao, Qianqian Wang, Shangzhan Zhang, Nan Xue, Sida Peng, Yujun Shen, and Xiaowei Zhou.
\newblock Spatialtracker: Tracking any 2d pixels in 3d space.
\newblock In \emph{Proceedings of the IEEE/CVF Conference on Computer Vision and Pattern Recognition}, pages 20406--20417, 2024.

\bibitem[Xie et~al.(2021)Xie, Wang, Yu, Anandkumar, Alvarez, and Luo]{xie2021segformer}
Enze Xie, Wenhai Wang, Zhiding Yu, Anima Anandkumar, Jose~M Alvarez, and Ping Luo.
\newblock Segformer: Simple and efficient design for semantic segmentation with transformers.
\newblock \emph{Advances in neural information processing systems}, 34:\penalty0 12077--12090, 2021.

\bibitem[Xie et~al.(2022)Xie, Wang, Wang, Cao, Yang, and Zeng]{xie2022correlation}
Fei Xie, Chunyu Wang, Guangting Wang, Yue Cao, Wankou Yang, and Wenjun Zeng.
\newblock Correlation-aware deep tracking.
\newblock In \emph{Proceedings of the IEEE/CVF conference on computer vision and pattern recognition}, pages 8751--8760, 2022.

\bibitem[Xu and Tao(2020)]{cider}
Qingshan Xu and Wenbing Tao.
\newblock Learning inverse depth regression for multi-view stereo with correlation cost volume.
\newblock In \emph{AAAI}, 2020.

\bibitem[Xu et~al.(2024)Xu, Shi, Yifan, Chen, Yang, Peng, Shen, and Wetzstein]{xu24grm:}
Yinghao Xu, Zifan Shi, Wang Yifan, Hansheng Chen, Ceyuan Yang, Sida Peng, Yujun Shen, and Gordon Wetzstein.
\newblock {GRM}: Large gaussian reconstruction model for efficient {3D} reconstruction and generation.
\newblock \emph{arXiv}, 2403.14621, 2024.

\bibitem[Yang et~al.(2025)Yang, Sax, Liang, Henaff, Tang, Cao, Chai, Meier, and Feiszli]{yang2025fast3r}
Jianing Yang, Alexander Sax, Kevin~J Liang, Mikael Henaff, Hao Tang, Ang Cao, Joyce Chai, Franziska Meier, and Matt Feiszli.
\newblock Fast3r: Towards 3d reconstruction of 1000+ images in one forward pass.
\newblock \emph{arXiv preprint arXiv:2501.13928}, 2025.

\bibitem[Yang et~al.(2024{\natexlab{a}})Yang, Kang, Huang, Xu, Feng, and Zhao]{yang24depth}
Lihe Yang, Bingyi Kang, Zilong Huang, Xiaogang Xu, Jiashi Feng, and Hengshuang Zhao.
\newblock Depth anything: Unleashing the power of large-scale unlabeled data.
\newblock In \emph{Proc. {CVPR}}, 2024{\natexlab{a}}.

\bibitem[Yang et~al.(2024{\natexlab{b}})Yang, Kang, Huang, Zhao, Xu, Feng, and Zhao]{depth_anything_v2}
Lihe Yang, Bingyi Kang, Zilong Huang, Zhen Zhao, Xiaogang Xu, Jiashi Feng, and Hengshuang Zhao.
\newblock Depth anything v2.
\newblock \emph{arXiv:2406.09414}, 2024{\natexlab{b}}.

\bibitem[Yao et~al.(2018{\natexlab{a}})Yao, Luo, Li, Fang, and Quan]{mvsnet}
Yao Yao, Zixin Luo, Shiwei Li, Tian Fang, and Long Quan.
\newblock Mvsnet: Depth inference for unstructured multi-view stereo.
\newblock In \emph{ECCV}, 2018{\natexlab{a}}.

\bibitem[Yao et~al.(2018{\natexlab{b}})Yao, Luo, Li, Fang, and Quan]{yao2018mvsnet}
Yao Yao, Zixin Luo, Shiwei Li, Tian Fang, and Long Quan.
\newblock Mvsnet: Depth inference for unstructured multi-view stereo.
\newblock In \emph{Proceedings of the European conference on computer vision (ECCV)}, pages 767--783, 2018{\natexlab{b}}.

\bibitem[Yao et~al.(2020)Yao, Luo, Li, Zhang, Ren, Zhou, Fang, and Quan]{yao2020blendedmvs}
Yao Yao, Zixin Luo, Shiwei Li, Jingyang Zhang, Yufan Ren, Lei Zhou, Tian Fang, and Long Quan.
\newblock Blendedmvs: A large-scale dataset for generalized multi-view stereo networks.
\newblock In \emph{Proceedings of the IEEE/CVF conference on computer vision and pattern recognition}, pages 1790--1799, 2020.

\bibitem[Yariv et~al.(2020)Yariv, Kasten, Moran, Galun, Atzmon, Ronen, and Lipman]{yariv2020multiview}
Lior Yariv, Yoni Kasten, Dror Moran, Meirav Galun, Matan Atzmon, Basri Ronen, and Yaron Lipman.
\newblock Multiview neural surface reconstruction by disentangling geometry and appearance.
\newblock \emph{Advances in Neural Information Processing Systems}, 33:\penalty0 2492--2502, 2020.

\bibitem[Yenduri et~al.(2023)Yenduri, M, G., Y, Srivastava, Maddikunta, G, Jhaveri, B, Wang, Vasilakos, and Gadekallu]{yenduri23generative}
Gokul Yenduri, Ramalingam M, Chemmalar~Selvi G., Supriya Y, Gautam Srivastava, Praveen Kumar~Reddy Maddikunta, Deepti~Raj G, Rutvij~H. Jhaveri, Prabadevi B, Weizheng Wang, Athanasios~V. Vasilakos, and Thippa~Reddy Gadekallu.
\newblock Generative pre-trained transformer: {A} comprehensive review on enabling technologies, potential applications, emerging challenges, and future directions.
\newblock \emph{arXiv.cs}, abs/2305.10435, 2023.

\bibitem[Yi et~al.(2016)Yi, Trulls, Lepetit, and Fua]{yi_lift_2016}
Kwang~Moo Yi, Eduard Trulls, Vincent Lepetit, and Pascal Fua.
\newblock {LIFT}: {Learned} {Invariant} {Feature} {Transform}.
\newblock In \emph{Proc. {ECCV}}, 2016.

\bibitem[Zhai et~al.(2023)Zhai, Likhomanenko, Littwin, Busbridge, Ramapuram, Zhang, Gu, and Susskind]{zhai2023stabilizing}
Shuangfei Zhai, Tatiana Likhomanenko, Etai Littwin, Dan Busbridge, Jason Ramapuram, Yizhe Zhang, Jiatao Gu, and Joshua~M Susskind.
\newblock Stabilizing transformer training by preventing attention entropy collapse.
\newblock In \emph{International Conference on Machine Learning}, pages 40770--40803. PMLR, 2023.

\bibitem[Zhang et~al.(2024{\natexlab{a}})Zhang, Herrmann, Hur, Jampani, Darrell, Cole, Sun, and Yang]{zhang24monst3r:}
Junyi Zhang, Charles Herrmann, Junhwa Hur, Varun Jampani, Trevor Darrell, Forrester Cole, Deqing Sun, and Ming-Hsuan Yang.
\newblock {MonST3R:} a simple approach for estimating geometry in the presence of motion.
\newblock \emph{arXiv}, 2410.03825, 2024{\natexlab{a}}.

\bibitem[Zhang et~al.(2022)Zhang, Ramanan, and Tulsiani]{zhang2022relpose}
Jason~Y Zhang, Deva Ramanan, and Shubham Tulsiani.
\newblock Relpose: Predicting probabilistic relative rotation for single objects in the wild.
\newblock In \emph{ECCV}, pages 592--611. Springer, 2022.

\bibitem[Zhang et~al.(2024{\natexlab{b}})Zhang, Lin, Kumar, Yang, Ramanan, and Tulsiani]{raydiffusion}
Jason~Y Zhang, Amy Lin, Moneish Kumar, Tzu-Hsuan Yang, Deva Ramanan, and Shubham Tulsiani.
\newblock Cameras as rays: Pose estimation via ray diffusion.
\newblock In \emph{International Conference on Learning Representations (ICLR)}, 2024{\natexlab{b}}.

\bibitem[Zhang et~al.(2024{\natexlab{c}})Zhang, Bi, Tan, Xiangli, Zhao, Sunkavalli, and Xu]{zhang2024gs}
Kai Zhang, Sai Bi, Hao Tan, Yuanbo Xiangli, Nanxuan Zhao, Kalyan Sunkavalli, and Zexiang Xu.
\newblock Gs-lrm: Large reconstruction model for 3d gaussian splatting.
\newblock In \emph{European Conference on Computer Vision}, pages 1--19. Springer, 2024{\natexlab{c}}.

\bibitem[Zhang et~al.(2024{\natexlab{d}})Zhang, Bi, Tan, Xiangli, Zhao, Sunkavalli, and Xu]{zhang24gs-lrm:}
Kai Zhang, Sai Bi, Hao Tan, Yuanbo Xiangli, Nanxuan Zhao, Kalyan Sunkavalli, and Zexiang Xu.
\newblock {GS-LRM:} large reconstruction model for {3D} {Gaussian} splatting.
\newblock \emph{arXiv}, 2404.19702, 2024{\natexlab{d}}.

\bibitem[Zhang et~al.(2025)Zhang, Wang, Xu, Xue, Rupprecht, Zhou, Shen, and Wetzstein]{zhang2025flare}
Shangzhan Zhang, Jianyuan Wang, Yinghao Xu, Nan Xue, Christian Rupprecht, Xiaowei Zhou, Yujun Shen, and Gordon Wetzstein.
\newblock Flare: Feed-forward geometry, appearance and camera estimation from uncalibrated sparse views, 2025.

\bibitem[Zhang et~al.(2023)Zhang, Peng, Hu, and Wang]{geomvsnet}
Zhe Zhang, Rui Peng, Yuxi Hu, and Ronggang Wang.
\newblock Geomvsnet: Learning multi-view stereo with geometry perception.
\newblock In \emph{CVPR}, 2023.

\bibitem[Zhao et~al.(2023)Zhao, Wu, Chen, Chen, Xu, and Li]{zhao2023aliked}
Xiaoming Zhao, Xingming Wu, Weihai Chen, Peter~CY Chen, Qingsong Xu, and Zhengguo Li.
\newblock Aliked: A lighter keypoint and descriptor extraction network via deformable transformation.
\newblock \emph{IEEE Transactions on Instrumentation and Measurement}, 72:\penalty0 1--16, 2023.

\bibitem[Zheng et~al.(2023)Zheng, Harley, Shen, Wetzstein, and Guibas]{zheng2023point}
Yang Zheng, Adam~W. Harley, Bokui Shen, Gordon Wetzstein, and Leonidas~J. Guibas.
\newblock Pointodyssey: A large-scale synthetic dataset for long-term point tracking.
\newblock In \emph{ICCV}, 2023.

\bibitem[Zhou et~al.(2017)Zhou, Brown, Snavely, and Lowe]{zhou2017unsupervised}
Tinghui Zhou, Matthew Brown, Noah Snavely, and David~G Lowe.
\newblock Unsupervised learning of depth and ego-motion from video.
\newblock In \emph{Proceedings of the IEEE conference on computer vision and pattern recognition}, pages 1851--1858, 2017.

\bibitem[Zhou et~al.(2018)Zhou, Tucker, Flynn, Fyffe, and Snavely]{zhou2018stereo}
Tinghui Zhou, Richard Tucker, John Flynn, Graham Fyffe, and Noah Snavely.
\newblock Stereo magnification: Learning view synthesis using multiplane images.
\newblock \emph{arXiv preprint arXiv:1805.09817}, 2018.

\end{thebibliography}
}
\end{document}